%% file: emnlp2022.tex
\pdfoutput=1

\documentclass[11pt]{article}

\usepackage[]{EMNLP2022}

\usepackage{times}
\usepackage{latexsym}

\usepackage[T1]{fontenc}

\usepackage[utf8]{inputenc}

\usepackage{microtype}

\usepackage{inconsolata}

\usepackage{amssymb}
\usepackage{multirow}
\usepackage{graphicx}
\usepackage{booktabs}
\usepackage{tabularx}
\usepackage{CJKutf8}
\usepackage{makecell}
\usepackage{tikz}
\usepackage[normalem]{ulem}
\usepackage{tcolorbox}
\usepackage{amsmath}
\usetikzlibrary{cd}
\usepackage{bm}
\usepackage{arydshln}
\usepackage{xcolor}
\usepackage{algorithm}
\usepackage{algpseudocode}
\usepackage{bbding}

\title{FCGEC: Fine-Grained Corpus for Chinese Grammatical Error Correction}

\author{Lvxiaowei Xu$^{1}$, Jianwang Wu$^{1}$, Jiawei Peng$^{1}$, Jiayu Fu$^{2}$, Ming Cai$^{1*}$ \\ $^{1}$Department of Computer Science and Technology, Zhejiang University \\
$^{2}$Base Station Platform Software Development Dept, Huawei Co., Ltd \\
\texttt{$^1$\{xlxw, wujw, pengjw, cm\}@zju.edu.cn} \\
\texttt{$^2$fionafu0808@gmail.com}}

\newcommand{\blue}[1]{\textcolor{blue}{#1}}
\newcommand{\mybox}[1]{\tikz[baseline=(MeNode.base)]{\node[rounded corners, fill=gray!20](MeNode){#1};}}

\newcommand\blfootnote[1]{%
  \begingroup
  \renewcommand\thefootnote{}\footnote{#1}%
  \addtocounter{footnote}{-1}%
  \endgroup
}

\definecolor{c1}{HTML}{BC0119} 
\definecolor{c2}{HTML}{002357} 
\definecolor{c3}{HTML}{006955} 
\definecolor{c4}{HTML}{633251} 

\begin{document}
\maketitle
\input{Chapters/abstract}

\input{Chapters/introduction}

\input{Chapters/data-construction}

\input{Chapters/models}

\input{Chapters/experiments}

\input{Chapters/related-work}

\input{Chapters/conclusion}

\input{Chapters/limitations}

\input{Chapters/ethics}

\input{Chapters/acknowlegement}

\bibliography{anthology,custom}
\bibliographystyle{acl_natbib}

\appendix

\input{Chapters/appendix}

\end{document}

%% file: Chapters/abstract.tex
\begin{abstract}
Grammatical Error Correction (GEC) has been broadly applied in automatic correction and proofreading system recently. However, it is still immature in Chinese GEC due to limited high-quality data from native speakers in terms of category and scale. In this paper, we present FCGEC, a fine-grained corpus to detect, identify and correct the grammatical errors. FCGEC is a human-annotated corpus with multiple references, consisting of 41,340 sentences collected mainly from multi-choice questions in public school Chinese examinations. Furthermore, we propose a  Switch-Tagger-Generator (STG) baseline model to correct the grammatical errors in low-resource settings. Compared to other GEC benchmark models, experimental results illustrate that STG outperforms them on our FCGEC. However, there exists a significant gap between benchmark models and humans that encourages future models to bridge it. Our annotation corpus and codes are available at \url{https://github.com/xlxwalex/FCGEC}$^{\dagger}$.
\end{abstract}

%% file: Chapters/introduction.tex
\section{Introduction}

Grammatical error correction (GEC) is a complex task, aiming at detecting, identifying and correcting various grammatical errors in a given sentence. GEC has recently attracted more attention due to its ability to correct and proofread the text, which can serve a variety of industries such as education, media and publishing \cite{wang2021comprehensive}.\blfootnote{$^*$ Corresponding author.}\blfootnote{$^{\dagger}$ Online evaluation site: \url{https://codalab.lisn.upsaclay.fr/competitions/8020}.}

However, Chinese GEC (CGEC) is still confronted with the following three obstacles: (1) \textbf{Lack of data.} The major obstacle in CGEC is that the high-quality manually annotated data is limited compared to other languages \cite{dahlmeier2013building,napoles2017jfleg,rozovskaya2019grammar,bryant2019bea,flachs2020grammatical,trinh2021new}. There are only five publicly accessible datasets in CGEC: NLPCC18 \cite{zhao2018overview} , CGED \cite{rao2020overview}, CTC-Qua, YACLC \cite{wang2021yaclc} and MuCGEC \cite{zhang2022mucgec}. (2) \textbf{Data sources are non-native speakers.} The sentences in NLPCC18, CGED, YACLC and MuCGEC are all collected from Chinese as a Foreign Language (CFL) learner sources. However, massive errors from native speakers rarely arise in these sources. Therefore, the native speaker errors are more challenging with the inclusion of pragmatic data. Though CTC-Qua covers grammatical errors in native speakers, it has insufficient scale with 972 sentences. (3) \textbf{Limited multiple references.} For an erroneous sentence, there tends to be different correction methods. The sentences revised by the model may be correct, but different from the ground truth. This may cause unexpected performance degradation \cite{bryant2015far}. Besides, more references can offer various correction schemas enabling the model to accommodate more scenarios. Among CGEC, only MuCGEC and YACLC provide rich references.

To tackle aforementioned obstacles, we present FCGEC, a large-scale fine-grained GEC corpus with  multiple  references. The sentences in FCGEC are mainly collected from multi-choice questions in public school Chinese examinations. Therefore, our FCGEC is more challenging since it involves more pragmatic data in the examinations of native speakers. As for multiple references, we assign 2 to 4 annotators on each sentence, thus more references can be attained in this way. Moreover, we generate more references in the annotation process through techniques with synonym substitution.

In order to correct the grammatical errors, recent works are mostly based on two categories of benchmark models. \textbf{Sequence-to-sequence (Seq2Seq)} approaches regard GEC as a generation task that straightforward converts an erroneous sentence to the correct one \cite{yuan2016grammatical,zhao2020maskgec,fu2018youdao}. However, training such a generation model requires more computational resources due to the autoregressive decoder. Moreover, the generated style of Seq2Seq models is more arbitrary, which is not well applicable for GEC task. More recently, \textbf{sequence-to-edit (Seq2Edit)} approaches gain interests which take GEC as a token-level labeling task \cite{awasthi2019parallel,omelianchuk2020gector} via different edits, such as \emph{insert}, \emph{delete}, etc. Nevertheless, previous work falls short on altering the word order and correcting errors simultaneously with iterating.

To fill these gaps, we propose Switch-Tagger-Generator (STG) model as an effective baseline to correct grammatical errors in low-resource settings inspired by \citet{mallinson2020felix}. Our STG can be decomposed into three modules: \emph{Switch} module determines the permutation of characters while \emph{Tagger} module identifies the operation tags of each character in the sequence. Notably, benefiting from carefully designed compound tags, we eliminate the necessity for iteration. As for \emph{Generator} module, we adopt non-autoregressive approach to fill in the characters that do not appear in the source. 

In summary, our contributions are as follows:
\begin{enumerate}
    \item We present FCGEC, a large-scale fine-grained corpus with multiple references and more challenging errors for CGEC.
    \item We propose a STG model and then conduct experiments to compare with two categories of benchmark models (Seq2Seq and Seq2Edit). Experimental results illustrate that our STG model outperforms these models on FCGEC.
    \item We find a significant gap between human performance and benchmark models that encourage future models to bridge it.
\end{enumerate}

%% file: Chapters/data-construction.tex
\section{Corpus Construction}

\subsection{Data Collection}

We collect raw sentences mainly from two resources to obtain various Chinese grammatical error corpus from native speakers. \textbf{(1) Public examination websites.} We crawl the multi-choice grammatical error problems (More erroneous sentences than correct sentences) through public websites which contain exercises and exams designed by teachers and experts. These problems cover public school examinations for native students from elementary to high school. \textbf{(2) News aggregator sites.} To balance the quantity of erroneous sentences and correct sentences, we attain a diverse range of high quality sentences without grammatical error in news aggregator sites.

In total, we collect 54,026 raw sentences from these resources. After removing duplicated or incomplete sentences, there are 41,340 sentences in our FCGEC corpus. We describe in detail the data sources and data structures in Appendix~\ref{appendix:source} \&~\ref{appendix:structure}.

\subsection{Fine-grained Data Format}
\label{sec:data-format}

To facilitate model for grammatical error detection and correction, we designate three-tier hierarchical levels of golden labels in FCGEC as follows:

\paragraph{Detection Level.} As a preliminary procedure to correcting grammatical errors, we require the binary classification of a given sentence according to whether it contains grammatical errors or not.

\paragraph{Identification Level.} The labels in this level could be regarded as necessary for a multi-class categorization problem. As the examples shown in Table~\ref{tab:error-type}, we group grammatical errors into seven categories based on the error hierarchy. The definition of error types are as follows: \textbf{Incorrect Word Collocation (IWC)} is a word-level grammatical error in which the related words are combined in the improper pattern. \textbf{Component Missing (CM)} and \textbf{Component Redundancy (CR)} are also word-level errors that some components (e.g., subject and object) of the sentence are missing or redundant. \textbf{Structure Confusion (SC)} is a syntax-level grammatical error that combines two similar grammatical structures into a single incorrect one. \textbf{Incorrect Word Order (IWO)} covers grammatical errors in word-level and pragmatic-level. Compared to the previous work \cite{zhang2022mucgec}, we also take into account the errors that require logic, common sense on top of syntax (e.g., recursive relationships). \textbf{Illogical (ILL)} and \textbf{Ambiguity (AM)} are pragmatic errors. The former comprises contradictory statements, while the latter includes expressions with indeterminate meanings.

\paragraph{Correction Level.} In the correction level, we propose an \emph{operation-oriented} paradigm to construct GEC labels instead of the \emph{error-coded} or \emph{rewriting} paradigms utilized in previous works \cite{ng2014conll,sakaguchi2016reassessing}. In \emph{rewriting} paradigms, the annotators directly rewrite the raw sentences to the correct sentences without grammatical errors. However, it is difficult for annotators to rewrite in a consistent style, which leads to a drop in annotation quality. As for the \emph{error-coded} paradigm, the annotators may diverge in determining the boundaries of the erroneous spans, thus raising the complexity of the procedure.

In contrast, the \emph{operation-oriented} paradigm is on the basis of four fundamental correction operations : \emph{Insert}, \emph{Delete}, \emph{Modify} and \emph{Switch}. As an example shown in Figure~\ref{fig:eg-operation}, this paradigm is more compatible with the conventions of the annotator when correcting errors. Meanwhile, annotators only need to consider what operations are required to correct the sentences, instead of paying attention to the erroneous span (e.g., the selection of the words is left to post-processing for unified optimization). In addition, we have a large amount of correction prompts (explanations of grammatical error problems) developed by teachers and experts based on these four operations that can be utilized to accelerate annotation process.

\input{Figures/operation-oriented-cn}

\subsection{Annotation Procedure}
The annotators are asked to follow the given prompts to complete  the three levels of labeling. Notably, we allow annotators to add unlimited references to sentences with grammatical errors based on the four operations in error correction level.

In order to improve annotation efficiency, we have developed a visual online tool to support the annotation procedure. In addition, we applied pattern matching and rule-based scripts to automatically convert a large amount (72.3\%) of prompts into operation labels. We show the interface of our visual annotation tool in Appendix~\ref{appendix:annotool}.

As for annotation process, we hire 20 undergraduate students and 4 expert examiners to annotate and verify the GEC labels. We follow the annotation procedure in SuperGLUE \cite{wang2019superglue} that each annotator should work on test data first. After that, they can compare their labels with the gold ones. We encourage them to discuss their mistakes, questions and standards with other annotators and experts. To attain high-quality annotation with multiple references, we duplicate the sentences in our corpus 2 to 4 times. Furthermore, it is guaranteed that the redundant sentences are annotated by different annotators. Then experts are asked to review data that the annotators could not in agreement on the labels and add reasonable references. It is worth mentioning that we search for possible synonyms of the characters generated by \emph{Insert} and \emph{Modify} operations in annotation. We believe that supplying more word choices to annotators can improve the multi-reference rate. Moreover, we set up a weekly communication meeting to discuss common issues in annotation and adapt the labeling criteria. In total, the entire annotation procedure lasted for more than 4 months.

\input{Tables/error-types}

\input{Tables/cgec-compare}

\input{Tables/statistics}

\subsection{Quality Control}
\label{sec:data-quality}

To ensure the high-quality of our FCGEC, we adopt the following five criteria: (1) Each sentence is inspected by two specialized annotators to correct spelling and punctuation errors before annotation. Meanwhile, they have to eliminate the incomplete sentences (due to unexpected text truncation). (2) The specialized annotators were also asked to tag the sentences from news aggregator source that might have grammatical problems while checking spelling errors. Then these potential sentences will be discussed in weekly communication meeting. (3) We ask the annotators to read our guidelines and annotate twenty test instances. Then experts check their accuracy of the annotation. The annotators that meet the accuracy (90\%) could continue to label the official data. (4) We assign 2x to 4x annotators per sentence for the corpus. In case their annotations are different, the experts will determine the correct labels. After that, annotators can also learn from these mistakes to achieve self-improvement. (5) After annotation, we unify the annotated labels under the minimal operation criteria inspired by \citet{dahlmeier2012better} which applies fewer operations during correcting grammatical errors. More details about minimal operation algorithm is described in Appendix \ref{appendix:minimal}.

\subsection{Data Statistics and Comparison}

We compare our corpus with other Chinese grammatical error datasets in Table~\ref{tab:dataset-compare}. Moreover, the concrete statistics of FCGEC are shown in Table~\ref{tab:dataset-statistic} and Appendix~\ref{appendix:errortype}. We summarize the advantages of our FCGEC in the following three aspects:

\paragraph{Multiple References.}
As discussed in \citet{bryant2015far} and \citet{zhang2022mucgec}, the training and evaluation of GEC models can benefit from multiple references. In order to obtain more references, we ask the annotators to submit different reasonable operations for correcting errors. Meanwhile, we specifically provide several choices of synonyms for the generated text during annotation. We search for synonyms using both fine and coarse granularity. The fine-grained approach is to obtain synonyms from electronic dictionaries, while the coarse-grained way relies on similarity of the word vectors. It enhances the ratio of multiple references.

\paragraph{More Pragmatic Data.}

Pragmatic data involves errors in logic, common sense, ambiguity, etc. We increase the proportion of pragmatic data (Table~\ref{tab:appendix-errortype}) compared to other CGEC datasets, thus rendering the data more complex and challenging. Notably, we fix the ambiguity errors by providing references to correct them from different semantics. 

\input{Figures/type-corrs}

\paragraph{Effective Error Types.}
We assign more refined error types to the grammatical errors, and these types are closely related to the correction operations. As shown in Figure~\ref{fig:type-corr}, error types are always highly relevant to particular operations (e.g., CM and CR rely on \emph{Insert} and \emph{Delete} operations respectively). We believe that error types can be utilized as auxiliary data to improve the performance of the GEC models to correct grammatical errors.

%% file: Figures/operation-oriented-cn.tex
\begin{CJK}{UTF8}{gkai}
\begin{figure}[t]
	\centering
	
\begin{tcolorbox}[colback=gray!10,
                  colframe=gray!10,
                  arc=1mm, auto outer arc,
                  left=5pt,
                  top=1pt,
                  bottom=1pt
                  ]
    \begin{tabular}[c]{@{}l@{}}
    \small \quad \quad \quad \quad \quad \quad \quad  \thinspace \thinspace \thinspace \thinspace \thinspace \thinspace \thinspace \blue{\textcolor{c1}{\textbf[{Delete}]不}} \thinspace \thinspace \thinspace \thinspace \thinspace
    \blue{\textcolor{c2}{\textbf{[Modify}]浮现$\rightarrow$发生}}\\ 
	为了避免地震的悲剧\mybox{\textcolor{c1}{\sout{不}}}再\mybox{\textcolor{c2}{浮现}}，我们都\\
	\small To prevent the tragedy of the earthquake from \textcolor{c1}{\sout{not}} \thinspace \textcolor{c2}{emer-} \\ 
	\small \quad \quad \thinspace \thinspace \thinspace \thinspace \textcolor{c1}{[\textbf{Delete}] not} \thinspace \thinspace \thinspace \thinspace \textcolor{c2}{{[\textbf{Modify}] emerging $\rightarrow$ happening}}
	\\\rule[3pt]{7cm}{0.01em} \\
	\small \textcolor{c3}{{[\textbf{Switch}] 加固$\leftrightarrow$ 建造}} \thinspace \thinspace \thinspace \thinspace \textcolor{c4}{{$\downarrow$[\textbf{Insert}] 避难所}}
	\\
	应该\mybox{\textcolor{c3}{加固}}并\mybox{\textcolor{c3}{建造}}\textcolor{c4}{。}
	\\ \small \textcolor{c2}{ging} again, we should \textcolor{c3}{fortify} and \textcolor{c3}{build} \textcolor{c4}{.} \\
	\small \textcolor{c3}{{[\textbf{Switch}] fortify$\leftrightarrow$ build}} \quad \quad \quad \quad \thinspace \thinspace \thinspace \thinspace \thinspace \thinspace \thinspace \thinspace \thinspace \thinspace \textcolor{c4}{{$\uparrow$[\textbf{Insert}] shelters}}
	\end{tabular}
	
\end{tcolorbox}
	\caption{An example of \emph{operation-oriented} paradigm.}
	\label{fig:eg-operation}
\end{figure}
\end{CJK}

%% file: Tables/error-types.tex
\begin{CJK}{UTF8}{gkai}

\begin{table}[t]
\begin{tabular}{cm{6.05cm}}
\toprule
\textbf{Type} & \makecell[c]{\textbf{Example}}\\ \hline

\begin{tabular}[c]{@{}l@{}}  \textbf{IWC} \end{tabular}&\begin{tabular}[c]{@{}l@{}} \small 自己有双聪明能干的手，什么都能做出来。\\ \small You have smart hands to do everything.\\ \small (Tips: ``hands'' cannot be combined with ``smart'') \end{tabular}     \\ \hline

\begin{tabular}[c]{@{}l@{}}  \normalsize \textbf{CM} \end{tabular} &\begin{tabular}[c]{@{}l@{}}\small 绿色植物具有产生氧气。\\ \small Plants have (the ability) to produce oxygen. \\ \small (Tips: Lack of object ``the ability'' )\end{tabular}\\ \hline

\begin{tabular}[c]{@{}l@{}} \normalsize  \textbf{CR}  \end{tabular} &\begin{tabular}[c]{@{}l@{}}\small 我们已走了约十里左右的路程。\\ \small We had walked about 10 miles or so. \\ \small (Tips: ``about'' and ``or so'' are redundant)\end{tabular}  \\ \hline

\begin{tabular}[c]{@{}l@{}}     \normalsize \textbf{SC} \end{tabular} &\begin{tabular}[c]{@{}l@{}}\small 交通事故发生的原因是开车看手机造成的。\\ \small Traffic accidents are caused by (because) looking \\ \small at cell phones while driving.\\ \small (Tips: the structure of ``because'' and ``caused by'' \\ \small cannot appear together in one sentence) \end{tabular}  \\ \hline

\begin{tabular}[c]{@{}l@{}}   \textbf{IWO} \end{tabular} &\begin{tabular}[c]{@{}l@{}}\small 我改正并认识了自己的错误。\\ \small I corrected and realized my fault. \\ \small (Tips: realize the fault first and correct it later) \end{tabular}  \\ \hline

\begin{tabular}[c]{@{}l@{}}   \textbf{ILL} \end{tabular} &\begin{tabular}[c]{@{}l@{}}\small 我们应该防止事故不发生。\\ \small We should prevent accidents from not occurring. \\ \small (Tips: double negation causes illogical errors)\end{tabular}  \\ \hline
\begin{tabular}[c]{@{}l@{}}   \textbf{AM}  \end{tabular} &\begin{tabular}[c]{@{}l@{}}\small 刚一开门，看病的就进来了。\\ \small As the door  opened, the doctor/patient came in.\\ \small (Tips: there is an ambiguity about who comes in)\end{tabular}  \\ 
\bottomrule 
\end{tabular}

\caption{Examples of different types of errors.}

\label{tab:error-type}
\end{table}

\end{CJK}

%% file: Tables/cgec-compare.tex
\begin{table*}[ht]
\centering
\begin{tabular}{ccccccc}
\toprule
\textbf{Corpus}                         & \textbf{Source}    & \textbf{Paradigm}          & \textbf{Sentence}     & \textbf{\#Error}   & \textbf{\#Refs} &  \textbf{\#Length}                \\ \hline
NLPCC\citeyearpar{zhao2018overview}  & CFL                & \emph{Error-coded}         &2000                      & 1983(99.15\%)      & 1.1                & 29.7                            \\
CGED                                   & CFL                & \emph{Error-coded}       &30145                      & 25837(85.71\%) & 1.0              & 46.6              \\
CTC-Qua(2021)                          & Native             & \emph{Error-coded}         &972                     & 482(49.59\%)       & 1.0                 &  48.9                  \\
MuCGEC\citeyearpar{zhang2022mucgec}      & CFL                & \emph{Rewriting}           &7063                  & 6544(92.65\%)      & 2.3                & 38.5                   \\ \hline
\textbf{FCGEC (Ours)} & \textbf{Native}    & \textbf{\emph{Operation}}  &\textbf{41340}         & \textbf{22517(54.47\%)}     & \textbf{1.7}   &\textbf{53.1}                            \\ \bottomrule
\end{tabular}
\caption{The comparison of different Chinese grammatical error correction corpus. Numbers in row \textbf{\#Error} mean the percentage of incorrect sentences in the corpus. \textbf{\#Refs} indicates the average number of references contained in each sentence on average while \textbf{\#Length} stands for the average number of characters in each sentence. Note that CGED is a combined corpus from 2016 to 2018 \cite{rao2018overview, rao2020overview}.}
\label{tab:dataset-compare}
\end{table*}

%% file: Tables/statistics.tex
\begin{table}[ht]
\fontsize{10}{12}\selectfont
\begin{tabular}{p{0.9cm} p{0.65cm} p{0.65cm} p{0.65cm} p{0.65cm} p{0.65cm} p{0.65cm}}
\toprule
\textbf{Subset} & \textbf{Sent.} & \textbf{Err.} & \textbf{\#S} & \textbf{\#D} & \textbf{\#I} & \textbf{\#M}\\ \hline
\textbf{Train}  & 36340 &  19761 & 3930  & 10468 & 8705  & 7459     \\
\textbf{Valid}  & 2000  &  1102  & 262   & 465  & 553   & 453     \\
\textbf{Test}   & 3000  &  1654  & 421   & 1496  & 919   &746    \\
\bottomrule 
\end{tabular}

\caption{Some statistics of FCGEC, including the number of sentences, the number of erroneous sentences, and the number of four operations (\#S, \#D, \#I, \#M denote \emph{Switch}, \emph{Delete}, \emph{Insert}, \emph{Modify}, respectively). }

\label{tab:dataset-statistic}
\end{table}

%% file: Figures/type-corrs.tex
\begin{figure}[t]
	\centering
	\includegraphics[width=\columnwidth]{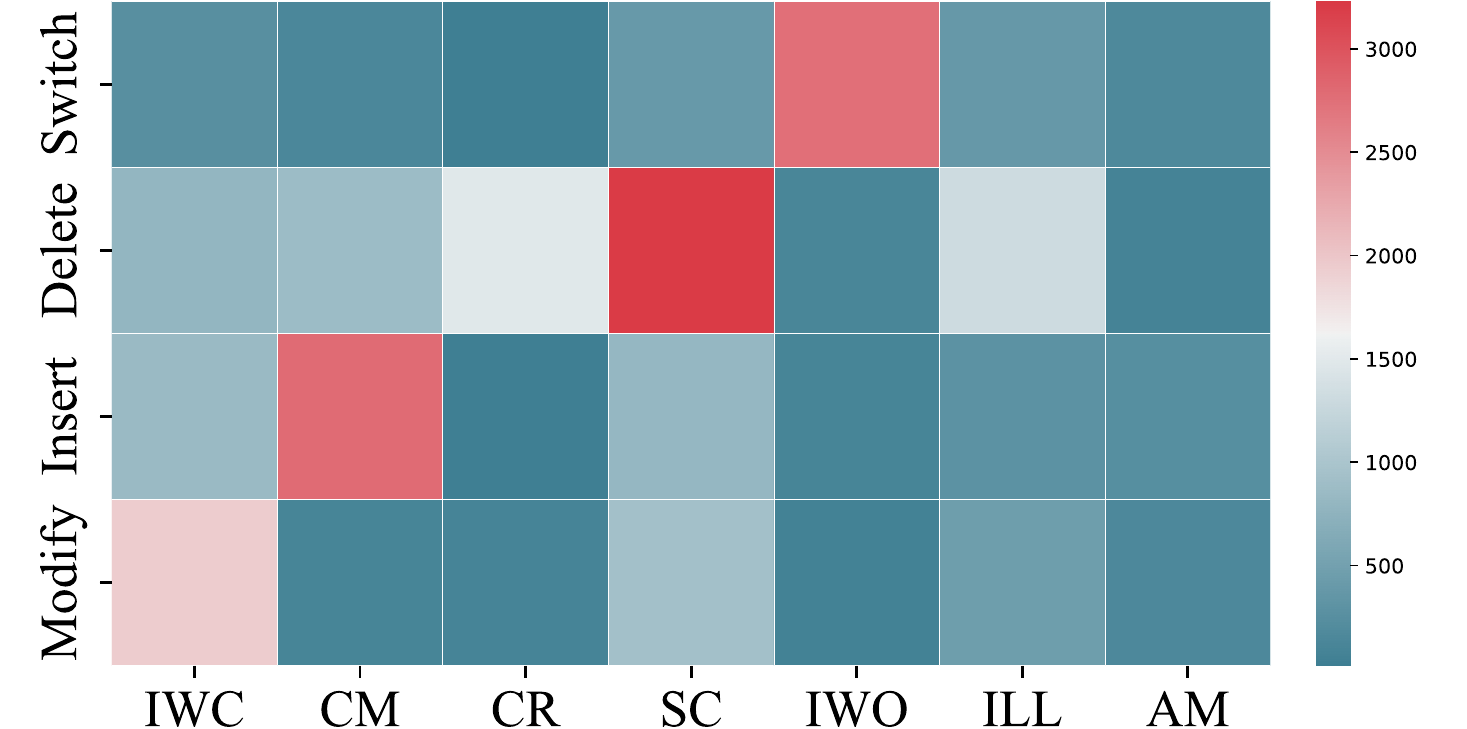} 
	\caption{Correlation between types and operations.}
	\label{fig:type-corr}
\end{figure}

%% file: Chapters/models.tex
\section{Benchmark Models}

We divide the GEC task into a classification task and a correction task. The classification task involves the error detection and error type identification. We adopt the pre-trained language models (PLMs) based approaches for these tasks. As for the correction task, we propose a STG model to correct errors. Meanwhile, two categories of mainstream GEC models (Seq2Seq and Seq2Edit) are applied as benchmark models for our FCGEC.

\subsection{Baselines for Classification Task}

In error detection subtask, the model needs to determine whether a given sentence contains grammatical errors. Therefore, it is a binary classification task while the type identification subtask can be regarded as a multi-class classification task. The model should predict which of the seven error types the given erroneous sentences belong to. Note that some sentences may have multiple error types.

Recently, PLMs are proved to be effective and achieve success in various fields, such as BERT \cite{devlin2018bert}, RoBERTa \cite{liu2019roberta}, BERT-WWM \cite{cui2021pre}, MacBERT \cite{cui2020revisiting} and StructBERT \cite{wang2019structbert}. Therefore, we adopt different PLMs enhanced models as the benchmark models for these classification tasks. Specifically, we treat multiple PLMs as the backbone network and apply fully-connected layers on top of it for detection and identification.

\subsection{Proposed STG Model}
\label{sec:stg-model}

To correct grammatical errors, we propose an effective benchmark model, STG, which tackles error correction in the low-resource settings. Figure~\ref{fig:stg-model} gives an overview of our model. STG decomposes the error correction task into three processing modules: \emph{Switch} , \emph{Tagger} and \emph{Generator}. The \emph{Switch} module determines the order of characters that appear in the output on the basis of pointer network \cite{vinyals2015pointer}. Our \emph{Tagger} module predicts the operation tags of each character and the number of characters that need to be generated in sequence. As for the \emph{Generator} module, it fills in the characters that are not present in source sentence. Notably, each module can be trained independently.

\paragraph{Switch Module.} The input to \emph{Switch} module for character $i$ is the hidden representation $h_i \in \mathbb{R}^{d}$ from PLM, where $d$ denotes the dimension of hidden representations. Then \emph{Switch} module determines the next position index for character $i$ based on the pointer network. We apply the self-attention to predict which possible character $e(i)$ would be pointed to. It can be formulated as:
\begin{equation}
    p\left(e(i)|h_i\right) = \operatorname{attention}\left(h_i, h_{e(i)} \right)
    \label{eq:1}
\end{equation}

The self-attention with scaled dot-product can be computed as below:
\begin{equation}
    A = \operatorname{Attention}\left(\bm{Q}, \bm{K}\right)= \operatorname{softmax}\left(\frac{\bm{Q}\bm{K}^{\mathrm{T}}}{\sqrt{d}}\right)
    \label{eq:2}
\end{equation}
where $A$ is attention score matrices, $\bm{Q}$ and $\bm{K}$ are both linear projections of $h$. More details about our \emph{Switch} module can be found in \ref{appendix:switch}. 

\input{Figures/stg-model}

\paragraph{Tagger Module.} We first define five tags corresponding to the three operations (except \emph{Switch} operation) as follows: the \emph{KEEP} (\emph{K}) tag is utilized to maintain the source character while the \emph{DEL} (\emph{D}) tag is assigned to remove character from source sequence. The tag of \emph{INS\_$t$} (\emph{I}\_$t$) represents the insertion of $t$ words after the current character. The substituted character is marked as \emph{MOD} (\emph{M}) tag for \emph{Modify} operation. As for the special case where the character is both substituted and required to insert other characters, we set the tag of \emph{MINS\_$t$} (\emph{MI}\_$t$) similar to \emph{I}\_$t$. Modification tags can perform a combination of multiple operations on a single character at the same time, thus eliminating the need to correct the sentence via iterations as other edit-based methods. Limited by space, we provide some concrete examples in the Appendix~\ref{appendix:tagger}.

We take the prediction of tags and the number $t$ of characters to be inserted or substituted for each character as classification task. Therefore, we apply two fully-connected layers to obtain tags and the number $t$. They can be written as:
\begin{equation}
    T = \sigma \left( Wh_i^s + b \right) 
    \label{eq:3}
\end{equation}
where $T$ denotes the tag or number $t$ while $h_i^s$ is the hidden representations of character $i$. And $\sigma$ stands for the $\operatorname{softmax}$ function. $W$ and $b$ are the learned weights and bias.

\paragraph{Generator Module.} As we can leverage the masked language modeling (MLM) task \cite{devlin2018bert} of BERT-style PLMs for generating the characters which do not appear in the source sequence, \emph{Generator} module inserts or substitutes the character with a certain number $t$ of \texttt{[MASK]} token according to their tags. Then it predicts which characters are suitable to fill into the masked places.

\paragraph{Training and Testing.} During the training process, we utilize cross-entropy loss $\mathcal{L}_{switch}$, $\mathcal{L}_{tag}$ and $\mathcal{L}_{gen}$ for the three modules. The STG model can be trained in two paradigms: \emph{independent} and \emph{joint}. The difference between them is whether each module is trained separately and thus they cannot utilize the shared encoder. We combine the loss in \emph{joint} paradigm as follows:
\begin{equation}
    \mathcal{L}(\theta) = \lambda_1 \mathcal{L}_{switch} + \lambda_2 \mathcal{L}_{tag} + \lambda_3 \mathcal{L}_{gen} + \gamma \lambda\|\Theta\|_{2}
    \label{eq:4}
\end{equation}
where $\lambda_\cdot$ is the coupling co-efficiency that regulates the three losses.$\Theta$ represents all trainable parameters in STG model and $\gamma$ denotes the coefficient of $L_2$-regularization. $\mathcal{L}_{tag}$ is always larger than the other two losses, thus we generally set it one order of magnitude smaller. Furthermore, we train STG model with type identification (TTI) task to utilize auxiliary type data to improve model performance.

As for testing phase, we feed the erroneous sentence into each module in a pipeline fashion to correct errors. Specifically, we adopt constrained beam search to decode the sequence order.

\subsection{Other Baselines for Correction Task}
\label{sec:benchmrk-gec}

In order to present mainstream error correction models on our corpus, we take two categories of approaches as benchmark models:

\paragraph{Seq2Seq Models.} A portion of the works adopt Transformer-based \cite{vaswani2017attention} encoder-decoder architecture as Seq2Seq fashion for correcting grammatical errors. The neural machine translation (NMT) based method is adopted in \citet{fu2018youdao} to tackle CGEC. Besides, \citet{kaneko2020encoder} utilize BERT-fuse to incorporate BERT into an encoder-decoder model for GEC. Meanwhile, MuCGEC \cite{zhang2022mucgec} presents a benchmark model based on Seq2Seq architecture with the Chinese BART \cite{shao2021cpt}.

\paragraph{Seq2Edit Models.} Recent works also focus on the Seq2Edit models, which correct errors by labeling manipulations of each character. LaserTagger \cite{malmi2019encode} is applied to modify the sequence with three types of edits: \emph{insertion}, \emph{deletion} and \emph{substitution}. PIE \cite{awasthi2019parallel} presents iterative edit with custom inflection operations to correct the grammatical errors. GECToR \cite{omelianchuk2020gector} is an iterative sequence tagging framework with custom g-transformations that we adapt it to CGEC follow the efforts of \citet{zhang2022mucgec}.

%% file: Figures/stg-model.tex
\begin{figure}[t]
	\centering
	\includegraphics[width=\columnwidth]{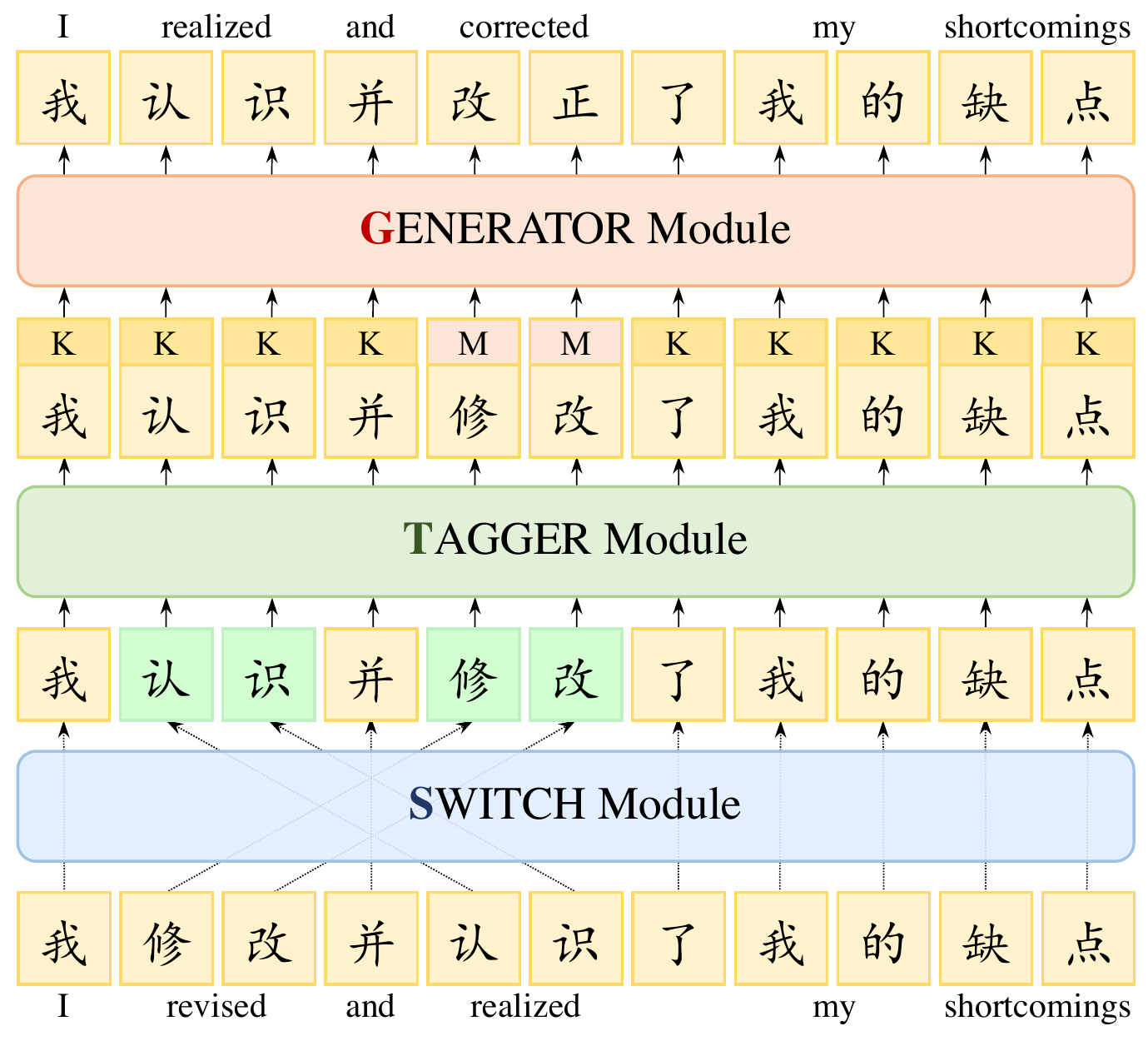} 
	\caption{The architecture of our STG model.}
	\label{fig:stg-model}
\end{figure}

%% file: Chapters/experiments.tex
\section{Experimental Results}

\subsection{Evaluation Metrics}
\paragraph{Classification Task.} We regard the error detection task and error type identification task as classification tasks. Therefore, we adopt four common metrics, i.e., \emph{Accuracy}, \emph{Precision}, \emph{Recall} and Macro \emph{F$_1$ score} to evaluate the model performance.

\paragraph{Correction Task.} As for correction task, we employ two different metrics : (1) \emph{Exact Match} metric is obtained by calculating the percentage of corrected sentences for model outputs that exactly matched with the golden references. (2) The character-level edit metrics proposed by MuCGEC \cite{zhang2022mucgec} are utilized to compute fine-grained model performance. After obtaining the optimal sequence for character-level editing, they merge consecutive edits of the same type into span-level for both model outputs and golden references. Then MuCGEC calculates the highest \emph{Precision}, \emph{Recall} and \emph{F$_{0.5}$ score} by comparing the edits of model outputs with each golden reference.

\subsection{Experimental Settings}

We conduct detailed experiments for fairly comparing benchmark approaches on our FCGEC. In classification tasks, we adopt officially released PLM parameters from HuggingFace website\footnote{\url{https://huggingface.co/models}}. Then we fine-tune the different PLMs on our FCGEC for 4 epochs with batch size of 64. As for the correction task, we employ RoBERTa as the backbone PLM of our STG model and other benchmark models for training 100 epochs. Notably, the BART PLM \cite{lewis2019bart} utilized in Seq2Seq models is substituted by CPT \cite{shao2021cpt}. We set maximum $t$ to 6 in \emph{Tagger} module (It can cover 98\% of the cases). In addition, we apply AdamW \cite{kingma2014adam} optimizer with a learning rate of 2e-5 and weight decay of 1e-2 for all tasks.

\subsection{Human Evaluation Results}

We hire 25 annotators from the crowd-sourcing platform of NetEase with a wide range of degrees and occupations. Specifically, annotators are restricted to be native speakers. We require them to annotate 10 instances for familiarization with the task requirements. Then they annotate 7,500 pieces of data (we randomly select 1,500 sentences from the test set and duplicate them 5 times), which is used to evaluate our human performance. As shown in Table~\ref{tab:overall-classify} and~\ref{tab:overall-correction}, the error detection task is relatively easy for humans while error type identification task is harder due to the fact that it has more categories. As for correction task, it is also challenging for humans to correct grammatical errors. We further discuss the human performance based on the model performance in Section~\ref{sec:overall-exp}.

\input{Tables/classification}
\subsection{Overall Performance}
\label{sec:overall-exp}

The results of the classification tasks on FCGEC are shown in Table~\ref{tab:overall-classify} for different PLMs from which several observations can be derived. First, the large-size variant PLMs perform better than other base-size models on the both detection and identification tasks as they can represent richer semantic information. To illustrate this observation, we roughly divide the error types into semantic and syntactic groups. We find that the average accuracy improvement for larger PLMs is significantly higher on the semantic group (7.6\%) compared to the syntactic group (2.8\%). Second, StructBERT-Large outperforms all PLMs on detection task while RoBERTa-Large achieves better performance on identification task, demonstrating two strong baselines at FCGEC. Moreover, there is an interesting observation on identification task that the humans have a lower performance of \emph{Recall} than all PLMs, while the \emph{Precision} is significantly better than them.

In terms of the correction task, Table~\ref{tab:overall-correction} demonstrates the results of benchmark models on FCGEC. The overall performances of Seq2Edit-based models are better than the Seq2Seq-based models. Furthermore, our STG-series models substantially outperform them on FCGEC, which proves the effectiveness of STG architecture. Finally, there is still a significant gap comparing best-performing models with humans in all tasks. Moreover, the difficulty of the task also increases gradually on the detection, identification and correction, which are reflected on the difference of gap between models and humans.

\subsection{Comparative Analysis}

\paragraph{Independent training vs. Joint training.} As we describe in Section~\ref{sec:stg-model}, the \emph{Switch}, \emph{Tagger}, \emph{Generator} modules in our STG can be trained flexibly either independently or jointly. In Table~\ref{tab:overall-correction}, STG with joint training (STG-Joint) brings gains of 4.17\% in EM score and 4.86\% in F$_{0.5}$ score compared with independent training STG (STG-Indep). The results indicate that the performance of correction can be enhanced during joint training since each module of STG can share more information and complement each other under a unified optimization objective. 

\paragraph{Investigate the benefit of error type identification to correction.} In Figure~\ref{fig:type-corr}, we illustrate the correlation between error types and the operations of correction. We observe a significant association among error types and operations, which motivates us to treat error type identification as an auxiliary task (TTI) for training STG model. As shown in Table~\ref{tab:overall-correction}, STG-Indep+TTI indicates that the three modules of STG are trained independently with the TTI task incorporated. Our STG achieves better performance after integrating the TTI task compared with STG-Indep, which demonstrates the efficient error type data can be utilized as auxiliary data to enhance model correction performance. Moreover, STG-Indep+TTI can also obtain an accuracy improvement of 1.78 points on the error type identification task compared to RoBERTa.

\input{Tables/correction}

\paragraph{Fine-grained performance analysis.} In Figure~\ref{fig:error-classify}, we demonstrate the fine-grained performance based on grammatical error types for identification task with RoBERTa-Large and correction task with STG-Joint. Notably, the dark sectors in the pie chart of the identification task indicate the proportion of errors for visual comparison. First, we observe the minimum error rate on SC, indicating that the PLM is more sensitive to syntactic structure errors. Second, the PLM performs weakly in terms of word-level errors, especially CR. After analyzing the error scenarios, we discover that the PLM may easily treat CR and CM errors as IWC errors. Furthermore, the PLM fails to determine the error types at the pragmatic level (i.e., ILL and AM), which illustrates the challenge of FCGEC.

As for correction task, it is clear that the performance on CM and IWC is inferior. We consider this potentially due to the fact that CM and IWC always require the generation of characters, increasing the difficulty of correction. Moreover, the model encounters trouble with AM due to the inclusion of pragmatic data such as ambiguity. It is hard for the model to distinguish the semantics in the sentences and correct it. In addition, we present more comparisons and fine-grained analyses in Appendix~\ref{appendix:more-analysis}.

\paragraph{Influence of the three modules in STG.} The correction performance of our STG model is affected by three modules simultaneously. Thus we further investigate the impact of these modules on the STG-Joint. As shown in Figure~\ref{fig:stg-sub-performance}, we analyze the char-level and sentence-level accuracy of each module. We ignore the \emph{Keep} tag when calculating the char-level accuracy in \emph{Tagger} module. \emph{Tagger-$t$ Acc.} is computed for the number $t$ of \emph{I}\_$t$ and \emph{MI}\_$t$ tags.  The first observation is that the performance of model is mainly constrained by \emph{Tagger} module, while it fails to predict tags and number $t$ precisely. Secondly, the performance of the \emph{Generator} module illustrates that it is possible to achieve acceptable performance via only utilizing non-autoregressive approach with fine-tuning. Furthermore, despite the high performance of the \emph{Switch} module, its role as the first module in the pipeline has a significant impact on the \emph{Tagger} and \emph{Generator} modules. Therefore a more robust performance of the \emph{Switch} module is needed.

%% file: Tables/classification.tex
\begin{table}[t]
    \fontsize{10}{12}\selectfont
	\centering
	\begin{tabular}{p{2.8cm} p{0.7cm}<{\centering} p{0.7cm}<{\centering} p{0.7cm}<{\centering}  p{0.7cm}<{\centering}}
		\toprule
	    \textbf{Model} & \textbf{Acc} & \textbf{P} & \textbf{R} & \textbf{F$_1$} \cr
		\midrule
		\multicolumn{5}{l}{$\bullet$ \textbf{Grammatical Error Detection}} \cr
		BERT               &72.17&71.99&72.12&71.75\cr 
		MacBERT            &74.12&74.15&74.25&73.98\cr 
		RoBERTa            &74.84&74.82&74.91&74.68\cr \hdashline 
		MacBERT-Large      &77.07&76.99&77.01&76.86\cr 
		RoBERTa-Large      &76.82&76.69&76.86&76.68\cr 
		StructBERT-Large   &\textbf{77.63}&\textbf{77.48}&\textbf{77.73}&\textbf{77.52}\cr \hdashline 
		Human              &\textbf{82.65}&\textbf{84.08}&\textbf{83.88}&\textbf{82.63}\cr \hline
		\multicolumn{5}{l}{$\bullet$ \textbf{Grammatical Error Type Identification}} \cr 
		BERT               &56.53&56.87&63.15&59.15\cr 
		MacBERT            &53.51&56.20&64.58&59.22\cr 
		RoBERTa            &54.90&58.78&65.34&61.33\cr \hdashline 
		MacBERT-Large      &57.86&57.29&63.51&59.76\cr 
		RoBERTa-Large      &\textbf{60.04}&59.86&\textbf{69.28}&\textbf{64.10}\cr 
		StructBERT-Large   &57.86&\textbf{60.13}&67.98&62.68\cr \hdashline 
		Human              &\textbf{73.04}&\textbf{76.74}&\textbf{57.15}&\textbf{64.40}\cr 
	\bottomrule
	\end{tabular}
	\caption{Average performance comparison on baselines among 10 independent runs for classification tasks.}
	\label{tab:overall-classify}
\end{table}

%% file: Tables/correction.tex
\begin{table}[t]
    \fontsize{10}{12}\selectfont
	\centering
	\begin{tabular}{p{2.8cm} p{0.7cm}<{\centering} p{0.7cm}<{\centering} p{0.7cm}<{\centering}  p{0.7cm}<{\centering}}
		\toprule
	    \textbf{Model} & \textbf{EM} & \textbf{P} & \textbf{R} & \textbf{F$_{0.5}$} \cr
		\midrule
		\multicolumn{5}{l}{$\bullet$ \textbf{Seq2Seq Models}} \cr
		BERT-fuse          &10.88&19.06&18.94&19.04\cr 
		MuCGEC             &21.16&39.47&26.19&35.84\cr  \hline
		\multicolumn{5}{l}{$\bullet$ \textbf{Seq2Edit Models}} \cr 
		PIE                &22.07&29.15&29.77&29.27\cr 
		LaserTagger        &28.42&36.60&31.16&35.36\cr 
		GECToR             &15.66&31.06&18.74&27.45\cr \hline
		\multicolumn{5}{l}{$\bullet$ \textbf{Our Models}} \cr
		STG-Indep          &29.93&43.16&32.88&40.62\cr 
		STG-Indep+TTI      &30.77&44.89&33.52&42.04\cr
		STG-Joint          &\textbf{34.10}&\textbf{48.19}&\textbf{37.14}&\textbf{45.48}\cr \hline
		Human              &\textbf{65.25}&\textbf{79.46}&\textbf{67.57}&\textbf{76.76}\cr 
	\bottomrule
	\end{tabular}
	\caption{Performance comparison for error correction tasks. Notably, EM indicates the metric of \emph{Exact Match}.}
	\label{tab:overall-correction}
\end{table}

%% file: Chapters/related-work.tex
\section{Related work}

\input{Figures/error-classify}

There already exists a lot of work on grammatical error correction for datasets and approaches. In terms of the dataset, most researches focus on English GEC. NUCLE \cite{dahlmeier2013building}, an early annotated corpus of GEC research, collects the erroneous sentences from students' essays in NUS. JFLEG \cite{napoles2017jfleg} is constructed from TOEFL exam with native sounding judgement. W\&I \cite{bryant2019bea} collects the texts from non-native English students around the world in an online web platform and then manually annotates them for GEC. By contrast, the errors in LOCNESS \cite{bryant2019bea} are acquired from essays written by native English students. Unlike the previous dataset, CWEB \cite{flachs2020grammatical} focuses on grammatical errors in low error density domains from websites. However, the scale of the data is relatively insufficient in CGEC. NLPCC \cite{zhao2018overview}, CGED \cite{rao2020overview}, YACLC \cite{wang2021yaclc} and MuCGEC \cite{zhang2022mucgec} are four publicly available non-native speaker resources for CGEC community, which encourage us to construct a high-quality CGEC corpus derived from native speakers.

As for the progress of GEC approaches, Seq2Seq and Seq2Edit are two mainstream approaches that achieve competitive results. Most of the work is based on Seq2Seq framework that generates the correct sentences directly \cite{zhou2019improving,wan2020improving,zhao2020maskgec,kaneko2020encoder}. Furthermore, after \citet{malmi2019encode} first apply the Seq2Edit approach, PIE \cite{awasthi2019parallel} and GECToR \cite{zhang2022mucgec} are proposed to correct errors with iterating. After that, \citet{tarnavskyi2022ensembling} employ an ensembling approach on GECToR for better performances. 

%% file: Figures/error-classify.tex
\begin{figure}[t]
	\centering
	\includegraphics[width=\columnwidth]{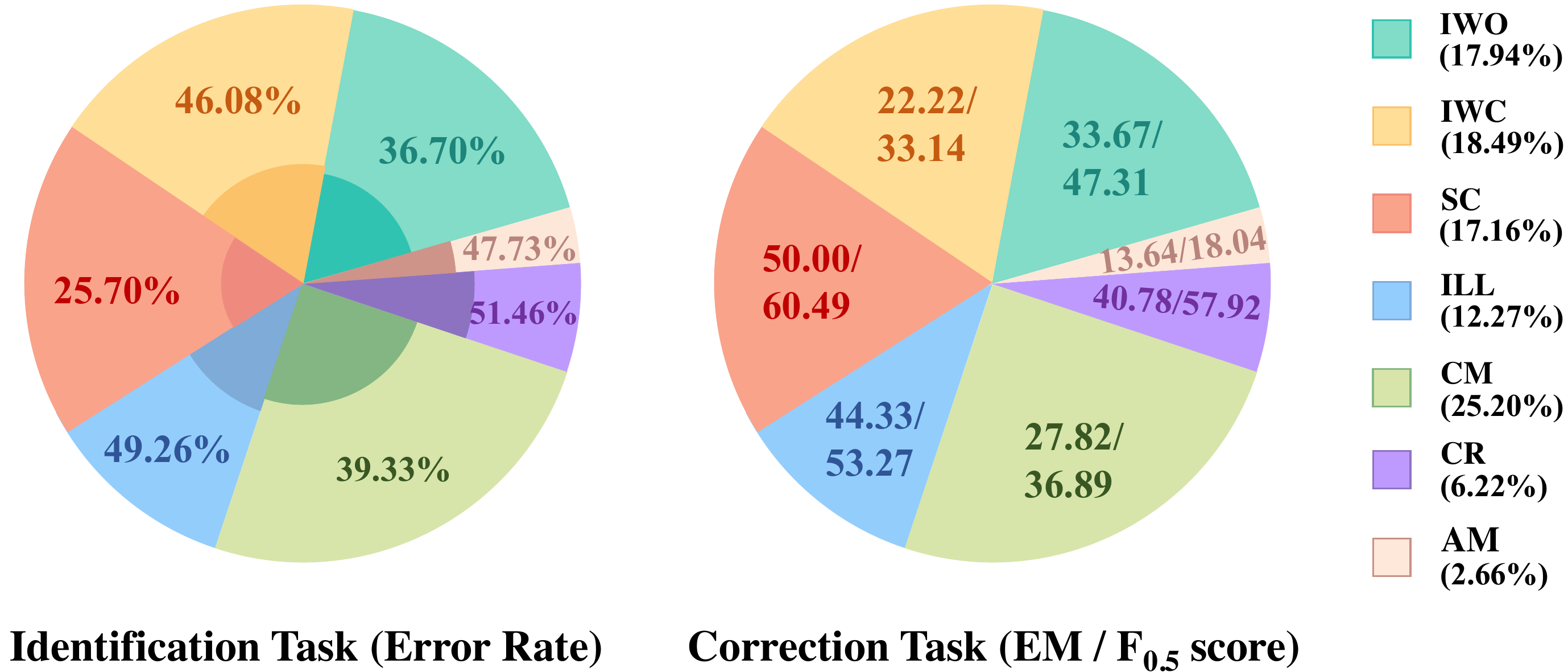} 
	\caption{The fine-grained performance on identification and correction tasks. The numbers in the sectors of the pie chart indicate the error rate and EM / F$_{0.5}$ score on identification task and correction task, respectively.}
	\label{fig:error-classify}
\end{figure}

%% file: Chapters/conclusion.tex
\section{Conclusion}

\input{Figures/stgsub-performance}

In this paper, we construct a large-scale corpus for Chinese grammatical error detection, identification and correction. Compared to previous CGEC corpus, our FCGEC is more complicated and challenging with pragmatic data. Furthermore, we provide multiple references so that the models can be evaluated for better performance. Furthermore, we propose a STG model to correct grammatical errors. Extensive experiments demonstrate that our STG outperforms the baselines and achieves the state-of-the-art performance. However, experiments show that there exists a notable gap between cutting-edge models and humans. Therefore, it encourages the future GEC models to bridge the gap.

%% file: Figures/stgsub-performance.tex
\begin{figure}[t]
	\centering
	\includegraphics[width=\columnwidth]{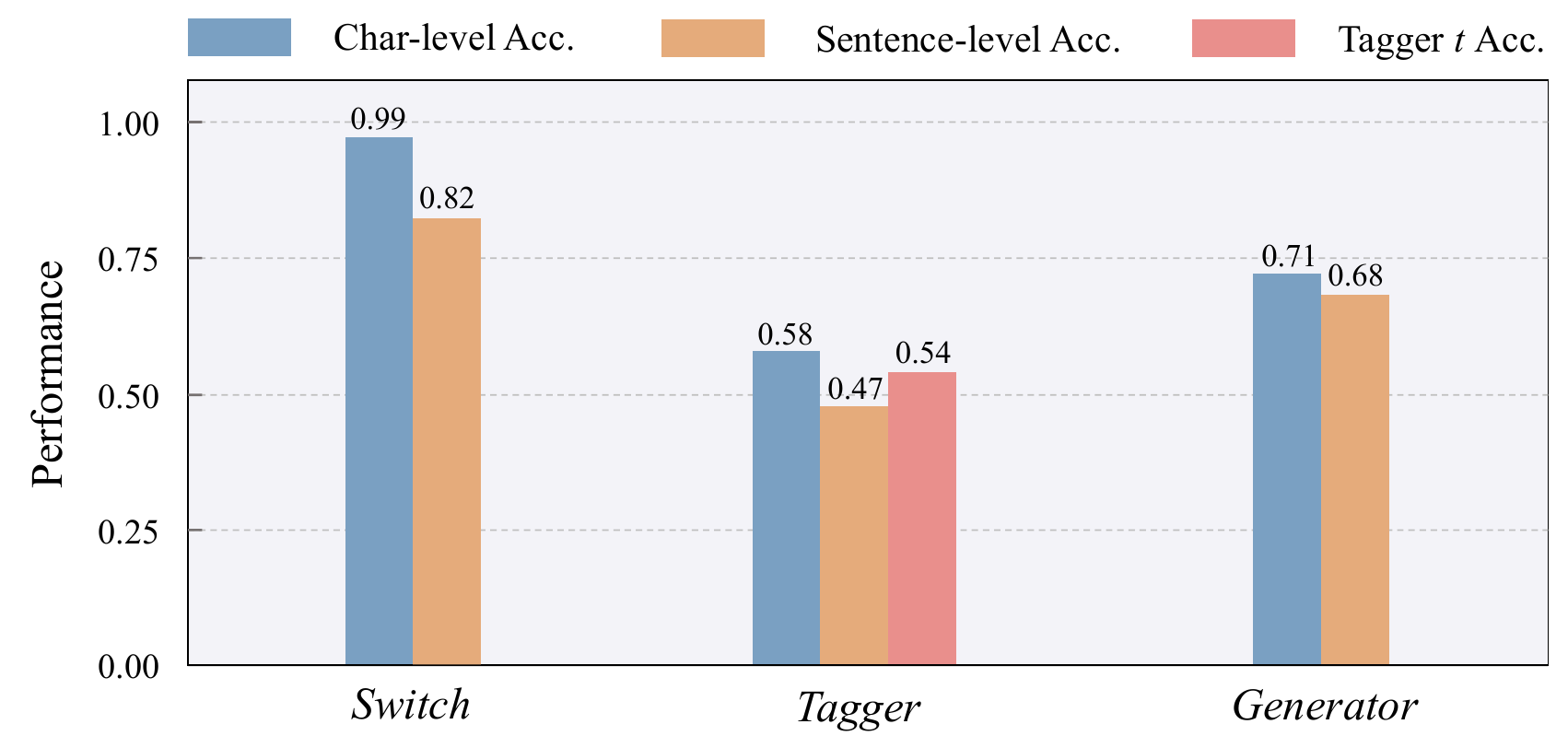} 
	\caption{The performances of three modules in STG.}
	\label{fig:stg-sub-performance}
\end{figure}

%% file: Chapters/limitations.tex
\section*{Limitations}
The limitations of our work can be categorized into two main aspects: our corpus and model.

\paragraph{Limitations of FCGEC.} In our FCGEC, a small number of sentences can be considered as different types of grammatical errors depending on correction methods. We do not provide a finer distinction between error types in this version. However, such fine-grained labels may supply more benefit in correction tasks (employing TTI as a auxiliary task). 

\paragraph{Limitations of STG.} The major limitation of our STG is that although no iteration is required, it corrects the errors via a pipeline paradigm with each modules in inference stage, thus it takes more time in the inference stage. Moreover, we consider that better performance may be achieved if the \emph{Generator} module is pre-trained with a massively parallel corpus such as Lang-8 \cite{zhao2018overview}, which we do not conduct in this paper.

%% file: Chapters/ethics.tex
\section*{Ethics Statement}

\paragraph{Licensing Issues.} FCGEC is a CGEC dataset collected from public examination websites and news aggregator sites. We collect the original grammatical error data or news data under the license of these sites or request for their permission. Meanwhile, the full attribution for original source of the data is cited in our FCGEC. In addition, we also commit not to use the corpus for commercial purposes, but only for the research studies.

\paragraph{Annotator Compensation.} In our annotation procedure, we hire two categories of annotators. The first type is the annotators who annotate or examine the data for our FCGEC. We estimate that a skillful annotator requires about 1 to 2 minutes for each sentence to identify the error types and correct errors. On this basis, we pay the annotator 7 yuan (about 1 dollar) for 10 sentences. The second category is annotators from the crowd-sourcing platform of NetEase for computing human performances on FCGEC. Since they only require to do the measurement of the data, we set the compensation to 4 yuan (about 0.6 dollar) per 10 sentences.

%% file: Chapters/acknowlegement.tex
\section*{Acknowledgements}
This research is supported by the Science and Technology Project of Zhejiang Province (2022C01044) and the National Natural Science Foundation of China (51775496).

%% file: Chapters/appendix.tex
\clearpage
\appendix

\section{Source of Data Collection}
\label{appendix:source}

For the source of public examination websites, we collect the practice exercises from KS5U (\url{http://5utk.ks5u.com/main.aspx}), which are accessible online for public usage. As for the news aggregator sites (e.g., ZAKER, IT Home etc.), we randomly collect the titles or topic sentences from these sites. After that, we manually check and correct the sentences as we describe in Section~\ref{sec:data-quality} to ensure the quality of our corpus.

\section{Data Structure}
\label{appendix:structure}

We employ the JSON format to construct our FCGEC, as illustrated below:
\input{Figures/data-format}

For the format of four operations, we define each of the operations as follows:

\input{Figures/data-operation}

\section{Visual Interface of Annotation Tool}
\label{appendix:annotool}

As shown in Figure~\ref{fig:anno-tool}, we develop a visual annotation tool for the annotation process. Given an erroneous sentence, annotators are able to utilize this tool for easily identifying the error types and correcting the errors via four operations, i.e., \emph{Switch}, \emph{Delete}, \emph{Insert} and \emph{Modify}. Concretely, the annotators only need to drag or click on the small blocks which represent each character at the bottom of the main form to complete the corresponding operation. In order to support multiple references of the correction task, we provide a button for adding a reference up to five. Moreover, the small form on the right side displays correction prompts from teachers and experts to inform the annotators of the correction. Our tool can automatically convert these prompts to operations via rules. In addition, the real-time correction labels are displayed on the bottom of this small form.

Furthermore, our tool enables convenient comparison of multiple annotation operations of a sentence by different annotators, so that expert examiners can select the reasonable references. In practice, this flexible annotation tool has greatly accelerated our annotation procedure.

\input{Figures/annotation_tool}

\section{More Statistics of FCGEC}
\label{appendix:errortype}

We conduct more statistics on our corpus, including the length distribution of sentences, the proportion of error types in each category and the distribution of the number of reference numbers.

\paragraph{Length distribution of sentences.}

We compute the distribution of sentence lengths in our corpus. As shown in Figure~\ref{fig:length-distrib}, the length of the longest sentence is 359, while the shortest only contains 9 characters. Furthermore, the average sentence length is 53.06. Based on this, we can use various types of PLMs to encode sentences from the corpus without having to deal with exceeding the length.

\input{Figures/length_distribute}

\paragraph{Distribution of error types.} In Table~\ref{tab:appendix-errortype}, we calculate the proportion of grammatical errors in the seven error categories on train set, validation set and test set. We can observe that  the error types are split as closely as possible with a similar distribution. Moreover, we can notice that the pragmatic type of error (ILL, AM and part of the IWO) also accounts for a significant proportion. Therefore our corpus tends to be more challenging.

\input{Tables/appendix-types}

\paragraph{The distribution of references.} For the erroneous sentences, we allow the annotators to correct them through a variety of references. Thus the distribution of sentences with respect to the number of references is shown in Figure~\ref{fig:multi-refernece}. First, we analyze sentences with a single reference and find that most of them are due to two scenarios: (1) The sentence is short or the grammatical error is very simple and obvious. (2) Some categories of errors often have only one way to correct them, such as IWO and ILL. Second, for sentences that contain two references, the errors of SC play a significant role in them. This is due to the fact that SC errors are always corrected by removing one of the two similar grammatical structures. More references for correcting sentences often indicate the need to insert or modify characters. Furthermore, we consider that the number of references could be increased via an enhanced and more refined annotation process and by assigning more annotators to each sentence.

\input{Figures/multiple-reference}
\input{Tables/demographic}

\section{Algorithm of Minimal Edit}
\label{appendix:minimal}

Given a pair containing the incorrect sentence and the corrected sentence, we design an algorithm to generate the operation labels under the minimal operation criteria from such pair. The algorithm is illustrated in Algorithm~\ref{algo:minimal}. It is mainly utilized in the following two scenarios: (1) Automatically convert the prompts of teachers and experts into operation labels for our visual annotation tool. (2) Based on this algorithm, we can unify the operation labels after annotation procedure to ensure that fewer operations are adopted during correcting grammatical errors for quality control. Besides, this can also help us to check for mistakes and guarantee the consistency of the data in the annotation.

In addition, we can convert the data from other formats (i.e., \emph{rewriting} and \emph{error-coded} paradigm) to our operation labels through this algorithm. Thus we can apply our STG model to other datasets.

\input{Figures/minimal-edit-operation}

\section{The Demographic of Humans}
\label{appendix:demographic}

In order to evaluate human performance on our FCGEC, we employ 25 annotators from the crowd-sourcing platform of NetEase. Moreover, with the aim of measuring human performance as completely as possible, we hire a diverse range of annotators with different aspects (Education, occupation, age, etc.). As shown in the Table~\ref{tab:demographic}, the platform of NetEase provides us with their non-private demographic information about the annotators.

It is worth mentioning that we ask the annotators to label more data (randomly sampling 50\% of the test set and then duplicating them 5 times) compared to other work as a way to attain more precise human performance.

\section{Details of Pre-trained Language Models}
\label{appendix:appendix-plm}

We enhance the performance of the model with PLMs for both the classification and correction tasks. In order to enable better reproduction of our results, we provide the details and links to officially released pre-trained parameters in the Table~\ref{tab:appendix-plms}.

In Seq2Seq models, we adopt CPT as the Chinese BART model. In particular, since some Chinese punctuation is missing in the vocabulary of the BART model (e.g., Chinese quotation marks), we avoid performance degradation by substituting the punctuation with their English counterparts during pre-processing stage.

\section{Details of Our STG Model}
\label{appendix:stg-appendix}

To better illustrate the details of our STG model, we present additional input samples and the processing for the three modules in this section.

\subsection{Switch Module}
\label{appendix:switch}
As we describe in Appendix~\ref{appendix:structure}, the labels of \emph{Switch} indicate the order of the original character index after swapping. However, the index of the next character is predicted for each character in our \emph{Switch} module. Therefore, we need to fill this gap by converting these labels. We demonstrate the differences between these two label types in Figure~\ref{fig:switch-label}.

\input{Figures/switch-label}

In \emph{Switch} module, we utilize pointer network with self-attention mechanism to predict which character will be pointed to of each character. Furthermore, we adopt cross-entropy as the loss function to measure the margin between attention score matrix $A$ and the golden labels for optimizing. 
\input{Tables/appendix-plms}

\subsection{Tagger Module}
\label{appendix:tagger}
In section~\ref{sec:stg-model}, we introduce five tags (\emph{K}, \emph{D}, \emph{INS\_$t$}, \emph{M} and \emph{MI}\_$t$) that correspond to the three operations (i.e., \emph{Delete}, \emph{Insert} and \emph{Modify}) in \emph{Tagger} module. For better understand our tags, we present some concrete examples in Figure~\ref{fig:tagger-label}.

\input{Figures/tagger-label}

With this well-designed tagging criterion, our STG model can perform arbitrary manipulations of the sequence without iteration. During the training stage, we employ two classification layers to determine the tags and the number $t$ of \emph{INS\_$t$} and \emph{MI}\_$t$, separately. Meanwhile, The cross-entropy is also applied to compute the loss. We optimize both of the parameters in the two classification layers simultaneously by combining the loss of them.

In particular, we try a small trick (Note that the trick is not adopted for the results of our STG models in Table~\ref{tab:overall-correction}, for fair comparison) that can further improve the performance of the model, which is to utilize the weighted cross-entropy loss. As the majority of tags in a sentence are \emph{Keep}, it is intuitive to increase the weight of other tags to solve this typical category imbalance problem. After we conduct experiment on STG-Joint with this trick, we observe a 0.72\% performance improvement in \emph{Exact Match}.

\input{Figures/generator-label}

\input{Tables/perform-operation}

\subsection{Generator Module}

In \emph{Generator} module, we exploit the features of BERT-style PLMs to predict the characters which do not appear in the source sentence. The outputs of \emph{Tagger} module are utilized to generate the input sequence with \texttt{[MASK]} tokens. We present an example of the input sequence in Figure~\ref{fig:generator-label}.

After that, the \emph{Generator} module predicts the indexes of the characters in vocabulary list that should be filled in at \texttt{[MASK]}. Similarly, the cross-entropy loss is adopted for optimizing the parameters in \emph{Generator} module.  

\section{More Comparisons and Analysis}
\label{appendix:more-analysis}

To further investigate the differences in the ability of the correction models, we present more comparisons and analyses in this section. 

\paragraph{Performance on four operations of correction.} Table~\ref{tab:perform-operate} illustrates another perspective of the fine-grained results that we calculate the performance of correction models on different operations. More specifically, we split the entire test set into four small subsets that contain only one operation for incorrect sentences. Then we compare our STG-Joint model with the best performing models in the Seq2Seq and Seq2Edit categories, respectively.

First, we can observe that our STG-Joint is significantly outperforms the other two models in terms of \emph{Switch} operation. This is due to the fact that we design a special \emph{Switch} module to efficiently handle such operation. Secondly, the performance of STG-Joint and LaserTagger is comparable in terms of \emph{Delete} operation. However, Seq2Seq model behaves relatively weakly, due to its arbitrary modifications that often tend to delete more characters. Lastly, we discover that all models have poor performances on \emph{Insert} and \emph{Modify} operations, indicating that the task of generating new characters is more challenging.

\paragraph{Original performance of MuCGEC.} In Section~\ref{sec:overall-exp}, we substitute the original backbone of the Seq2Seq and Seq2Edit models in MuCGEC \cite{zhang2022mucgec} for a fair comparison. We conduct additional experiments to demonstrate the original performance of models in MuCGEC. They employ the PLMs of Chinese-BART-Large\footnote{\url{https://huggingface.co/fnlp/bart-large-chinese}} and StructBERT-Large\footnote{\url{https://github.com/alibaba/AliceMind/tree/main/StructBERT}} for Seq2Seq and Seq2Edit (GECToR) model, respectively. 

We present the results of MuCGEC in Table~\ref{tab:appendix-mucegec}. It is clear that the performances of models with large-size PLMs are better than those of the base-size PLMs. Specifically, Seq2Seq model is greatly improved after applying BART-Large as the backbone. Moreover, it is close to the results of our STG-series models. However, there is only a slight improvement of \emph{F$_{0.5}$} for the Seq2Edit model. After we further observe the error examples of Seq2Seq and Seq2Edit on the test set, we discover that the results of \emph{Switch} operation are the critical limitation for Seq2Edit model. This also illustrates the necessity of the \emph{Switch} module in our STG.

\input{Tables/appendix-mucgec}

\paragraph{Performances on the validation set.} In order to evaluate the distribution of our split dataset, we show the results of the best performing models on the corresponding validation set for the three tasks (detection, identification and correction) in Table~\ref{tab:appendix-valid}. It is reasonable to observe that the performance on the validation set is slightly better than on the test set. Therefore, we keep the data distribution on the validation set close to the test set, which facilitates the model to search for the best hyperparameters on the validation set. Thus we can obtain better performance on the test set.

\input{Tables/appendix-valid}

\paragraph{Computing resources and times.} In Table~\ref{tab:appendix-compute}, we show the detailed computing resources and hyperparameters for training our STG-Joint model. Meanwhile, we also record the training time consumed under these hyperparameters and devices.

\input{Tables/appendix-computation}

\section{Case Study}
\label{appendix:case-study}

In order to explore the performance of the models on the correction task, we conduct analyses on case examples. Similarly, we compare STG-Joint with the best performing models in the Seq2Seq and Seq2Edit categories (MuCGEC and LaserTagger). We present the cases in Table~\ref{tab:case-study}. Meanwhile, the English version of the cases can be seen in Table~\ref{tab:case-study-eng}. In particular, since the ground truth contains multiple references, we represent one of them as an illustration due to space constraints. Note that the results of the models in Table~\ref{tab:case-study} are based on the multiple references.

We can derive several observations from these case examples. First, in the category of word order errors (IWO), both Seq2Seq and Seq2Edit models can correct the elementary errors (Example 10). However, they fail to solve the more difficult order errors (progressive relationship problem in Example 9). Since our STG model is specifically equipped with the \emph{Switch} module, it is possible to correct for these errors. Secondly, in the case of error categories that require the generation of new 
characters (e.g. CM and IWC), more improvements are required for all models. Finally, the pragmatic errors are the most difficult to correct (especially for AM). We encourage future models to pay more attention to these types of errors.

\input{Tables/case-study}

\input{Tables/case-study-eng}

%% file: Figures/data-format.tex
\begin{tcolorbox}[colback=white,
                  colframe=black,
                  arc=1mm, auto outer arc,
                  boxrule=0.5pt,
                  left=5pt,
                  top=1pt,
                  bottom=1pt
                  ]
                  
    \begin{tabular}[c]{@{}l@{}}
    \small \{ \\
    \small \quad ``id(The global id of the instance)'': \{ \\
    \small \quad \quad``sentence'': The original sentence, \\
    \small \quad \quad``error\_flag'':  Whether sentence contains errors, \\
    \small \quad \quad``error\_type'': The error types of sentence, \\  
    \small \quad \quad``operation'' : [ \\
    \small \quad \quad \quad\{ The operation of the first reference \}, \\
    \small \quad \quad \quad\{ The operation of the second reference \}, \\
    \small \quad \quad \quad\{ ... \}], \\
    \small \quad \quad``external'' : Additional information \\
    \small \quad\} \\
    \small \}
	\end{tabular}
\end{tcolorbox}

%% file: Figures/data-operation.tex
\begin{tcolorbox}[colback=white,
                  colframe=black,
                  arc=1mm, auto outer arc,
                  boxrule=0.5pt,
                  left=5pt,
                  top=1pt,
                  bottom=1pt
                  ]
                  
    \begin{tabular}[c]{@{}l@{}}
    \small >> Suppose the given sentence is ``A B C D E''.   \\
    \small \textbf{1. \emph{Switch} operation} \\
    \small \blue{\{``Switch'':[0,2,1,3,4]\}} \, (``A B C D E'' $\to$ ``A C B D E'') \\ 
    \small // The values in the list indicate the order of the original \\
    \small character index after the swap (Index starts from 0). \\
    \small \textbf{2. \emph{Delete} operation} \\
    \small \blue{\{``Delete'':[3]\}} \quad \quad \quad  \quad \, (``A B C D E'' $\to$ ``A B C E'') \\ 
    \small // The characters indexed in the list will be deleted. \\
    \small \textbf{3. \emph{Insert} operation} \\
    \small \blue{\{``Insert'':[\{``pos'':1,``tag'':``INS\_1'',``label'':[``F'']\}]\}} \\
    \small \quad \quad \quad \quad \quad \quad  \quad \quad \, \,  (``A B C D E'' $\to$ ``A B F C D E'') \\ 
    \small // Insert a ``F'' after the character indexed by ``pos''. \\
    \small \textbf{4. \emph{Modify} operation} \\
    \small \blue{\{``Modify'':[\{``pos'':2,``tag'':``MOD\_1'',``label'':[``F'']\}]\}} \\
    \small \quad \quad \quad \quad \quad \quad  \quad \quad \quad \, \,  (``A B C D E'' $\to$ ``A B F D E'') \\ 
    \small // Modify the character with index ``pos'' to ``F''.
	\end{tabular}
\end{tcolorbox}

%% file: Figures/annotation_tool.tex
\begin{figure}[ht]
	\centering
	\includegraphics[width=\columnwidth]{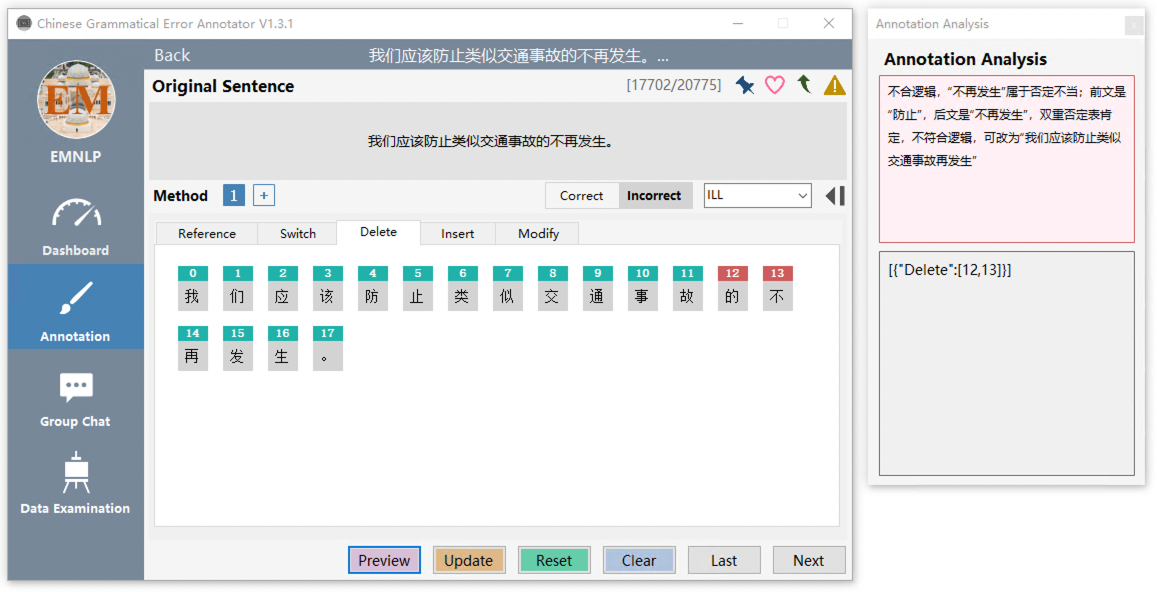} 
	\caption{The screenshot of our annotation tool.}
	\label{fig:anno-tool}
\end{figure}

%% file: Figures/length_distribute.tex
\begin{figure}[ht]
	\centering
	\includegraphics[width=\columnwidth]{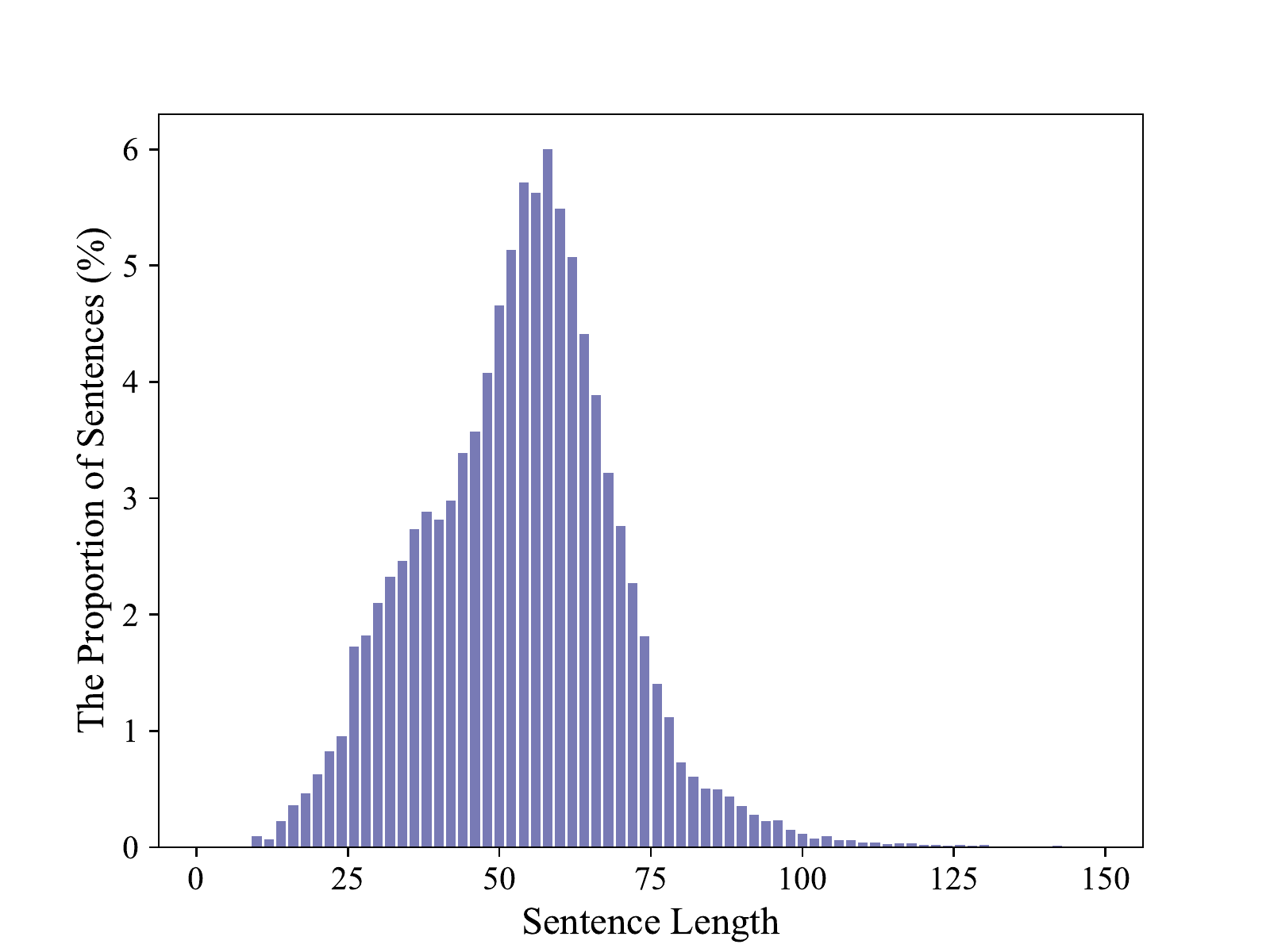} 
	\caption{The length distribution of sentences in the whole corpus.}
	\label{fig:length-distrib}
\end{figure}

%% file: Tables/appendix-types.tex
\begin{table}[t]
\fontsize{10}{12}\selectfont
\begin{tabular}{p{0.67cm} p{0.67cm} p{0.67cm} p{0.67cm} p{0.67cm} p{0.67cm} p{0.67cm} p{0.67cm}}
\toprule
\textbf{IWC} & \textbf{CM} & \textbf{CR} & \textbf{SC} & \textbf{IWO} & \textbf{ILL} & \textbf{AM} \\ \hline
\multicolumn{7}{l}{$\bullet$ \textbf{Train set}} \cr

19.45 &  19.83  & 8.07  & 15.93 & 14.25 & 17.56  & 4.92     \cr \hdashline 
\multicolumn{7}{l}{$\bullet$ \textbf{Validation set}}   \cr
21.58 &  20.71  & 6.41  & 19.41 & 16.90  & 11.27  & 3.73    \cr \hdashline 
\multicolumn{7}{l}{$\bullet$ \textbf{Test set}}   \cr
18.50  & 25.21  & 6.23  & 17.17 & 17.96  & 12.27  & 2.66    \cr
\bottomrule 
\end{tabular}

\caption{The proportion (\%) of error types in Train, Validation and Test set, respectively.}

\label{tab:appendix-errortype}
\end{table}

%% file: Figures/multiple-reference.tex
\begin{figure}[t]
	\centering
	\includegraphics[width=0.75\columnwidth]{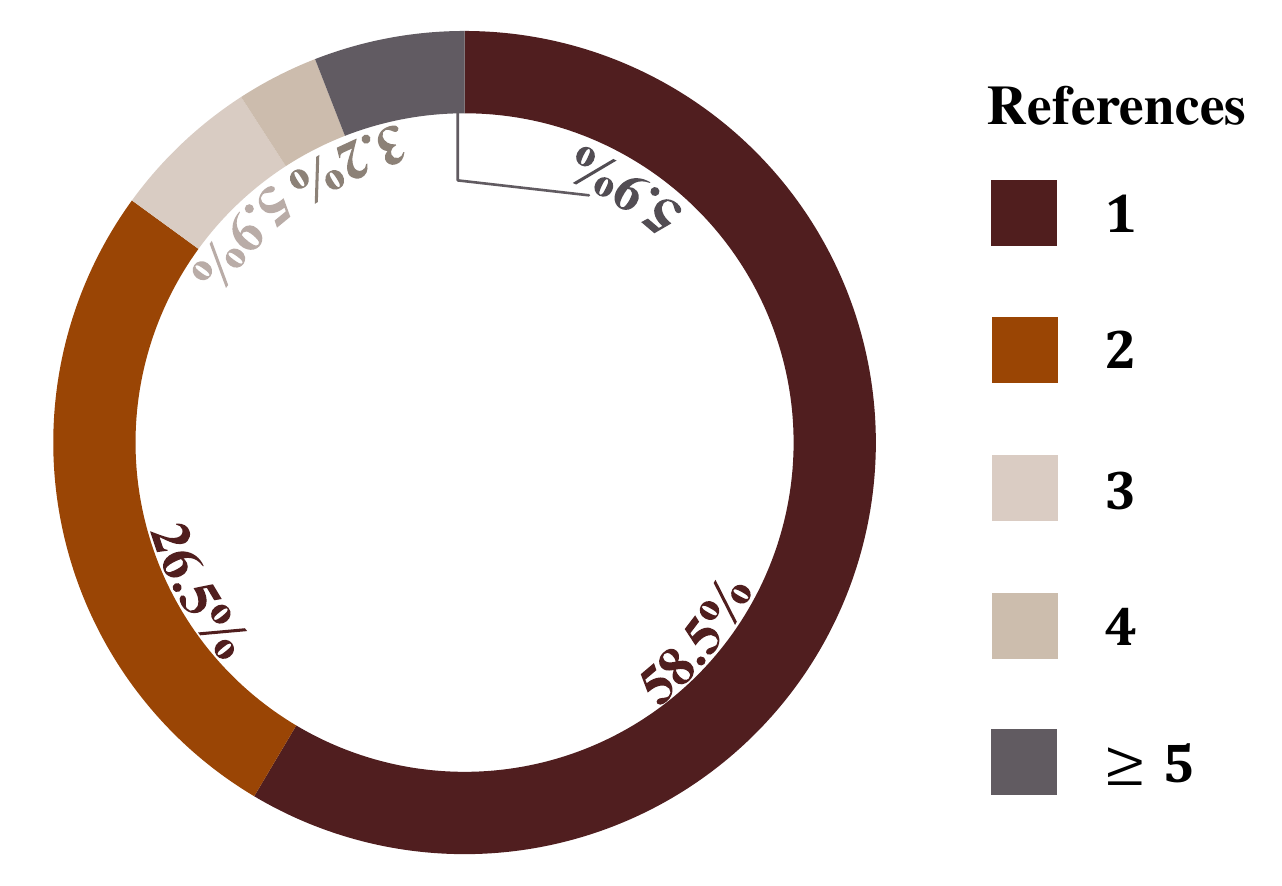} 
	\caption{The distribution of the sentences in terms of references in our FCGEC.}
	\label{fig:multi-refernece}
\end{figure}

%% file: Tables/demographic.tex
\begin{table}[t]
\fontsize{9}{12}\selectfont
\begin{tabular}{p{3cm} p{3.8cm}}
\toprule
\textbf{Attribute} & \textbf{Value} \\ \midrule
Gender (Male / Female)   & 9 / 16     \cr  
Age                      & [18, 54]      \cr 
Native Language          & Chinese  \cr \hline
Education                & \begin{tabular}[c]{@{}l@{}}  Middle School Student \\ Specialty \\ Undergraduate Student \\Bachelor's Degree \\  Master's Degree \end{tabular} \cr  \hline
Occupation               & \begin{tabular}[c]{@{}l@{}} Students \\ Freelancers \\ Factory Workers \\ Federal Employees \\ Professionals (e.g., Lawyers) \\ Office Staffs (e.g., Managers)\end{tabular} \cr
\bottomrule 
\end{tabular}

\caption{The demographic information of annotators. }

\label{tab:demographic}
\end{table}

%% file: Figures/minimal-edit-operation.tex
\makeatletter
\newcommand{\multiline}[1]{%
  \begin{tabularx}{\dimexpr\linewidth-\ALG@thistlm}[t]{@{}X@{}}
    #1
  \end{tabularx}
}
\makeatother

\begin{algorithm}[ht]
    \small
    \caption{Attain the operation labels via minimum edit distance.}
    \begin{algorithmic}[1]
        \Statex {\bf{ Input:}}
            Source sentence $S$ and target sentence $T$.
        \Statex {\bf{ Output:}}
            The operation labels $L$ that convert $S$ to $T$.
        \If{S == T}
            \State {\bf return} [\{\}]
        \EndIf
        \State $L = [\{\}]$
        \State $tags =  [$``Copy'', ``Modify'', ``Delete'', ``Insert''$]$
        \State $mov = [(-1, -1), (-1, -1), (-1, 0), (0, -1)]$
        \State Calculate the character frequency $f_S$ of $S$, and $f_T$ of $T$.
        \If{$f_S == f_T$}
            \State \multiline{%
            Calculate the longest common substring $s_1$ and second longest one $s_2$ between $S$ and $T$.} 
            \State \multiline{
            Swap the positions of $s_1$ and $s_2$ with their indexes $p_{ori}$ in $S$. Then the swapped sentence $S_{swap}$ and indexes $p_{swap}$ can be obtained.}
            \If{$S_{swap} == T$}
                \State $L = [\{\text{``Switch''}: p_{swap}\}]$
            \Else 
                \State {\bf goto} $17$
            \EndIf
        \Else
            \State \multiline{
            Calculate the Levenshtein distance and then obtain the edit path matrix $M_p$.}
            \State $ops = []$
            \State getOperations($M_p, \text{len}(S), \text{len}(T), ops$)
            \State \multiline{Merge labels with adjacent index and the same operation in $ops$ to get $L$}
        \EndIf
        \State {\bf return} $L$
        
        \;
        
        \State {\bf Function} getOperations($M_p, i, j, ops$)
            \If{$i == 0 \text{ and } j == 0$}
                \State {\bf return}
            \EndIf
            \ForAll {$op$ such that $op \in tags$ }
                \If{$M_p[i][j].get(op)$}
                    \State $ops$.append($[i, j, op]$)
                    \State \multiline{
                    getOperations($M_p, i + mov[0], j + mov[1], ops$)}
                    \State {\bf break}
                \EndIf
            \EndFor
        \State {\bf EndFunction}
    \end{algorithmic}    
    \label{algo:minimal}
\end{algorithm}

%% file: Figures/switch-label.tex
\begin{figure}[ht]
	\centering
	\includegraphics[width=\columnwidth]{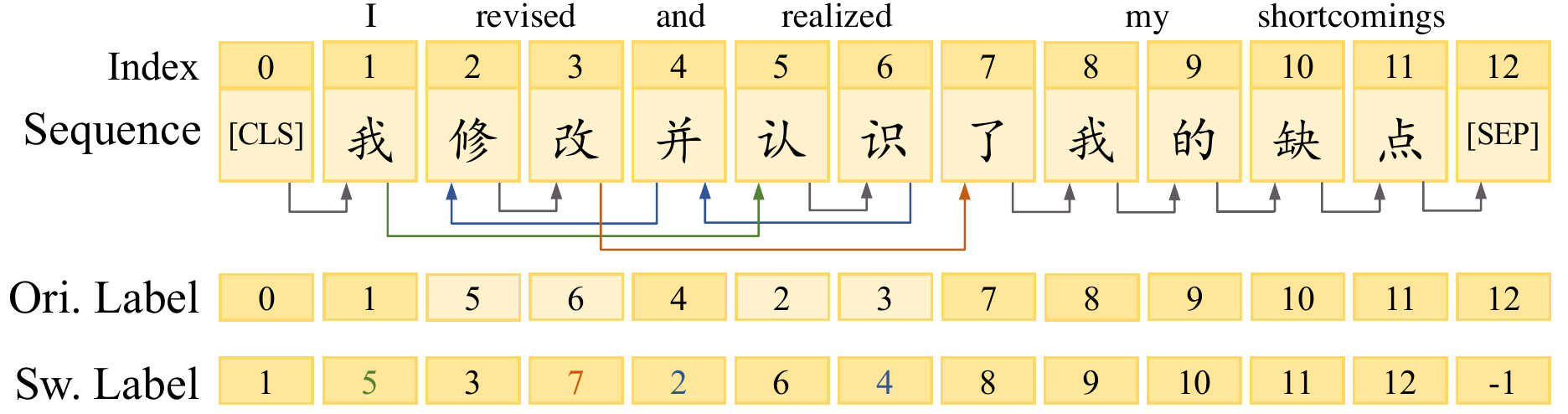} 
	\caption{Comparison of the differences in the two types of \emph{Switch} labels. \textbf{Ori. Label} and \textbf{Sw. Label} indicate the annotated labels and the processed labels in \emph{Switch} module, respectively.}
	\label{fig:switch-label}
\end{figure}

%% file: Tables/appendix-plms.tex
\begin{table*}[ht]
\centering
\fontsize{9}{12}\selectfont
\begin{tabular}{p{4cm} p{2.5cm} p{6cm} c}
\toprule
\textbf{PLM Name} & \textbf{Parameters} &\textbf{Url} & \textbf{Model Size}\\ \midrule
\multicolumn{4}{l}{\textbf{$\bullet$ PLMs of Basic Size}}\cr
\begin{tabular}[c]{@{}l@{}}BERT \cite{devlin2018bert} \\ \small $\to$ BERT-wwm-ext\end{tabular}    & \multirow{4}{*}{ \begin{tabular}[c]{@{}l@{}}12 layers \\768-d hidden Size \\12 attention heads \end{tabular}} &\multirow{2}{*}{\url{https://github.com/ymcui/Chinese-BERT-wwm}} & \multirow{4}{*}{102M}   \cr  
\begin{tabular}[c]{@{}l@{}}RoBERTa$^\dagger$ \cite{liu2019roberta} \\ \small $\to$ RoBERTa-wwm-ext \end{tabular}           & &  &   \cr 
\begin{tabular}[c]{@{}l@{}}MACBERT \cite{cui2020revisiting} \\ \small $\to$ MacBERT-base \end{tabular}      & & \url{https://github.com/ymcui/MacBERT} & \cr \hdashline
\begin{tabular}[c]{@{}l@{}}CPT$^\dagger$ \cite{shao2021cpt} \\ \small $\to$  CPT-base \end{tabular}      & \begin{tabular}[c]{@{}l@{}}10 layers of encoder \\ 2 layers of decoder \\768-d hidden Size \\12 attention heads \end{tabular} & \url{https://github.com/fastnlp/CPT} & 116M \cr \hline

\multicolumn{4}{l}{\textbf{$\bullet$ PLMs of Large Size}}\cr
\begin{tabular}[c]{@{}l@{}}RoBERTa \cite{liu2019roberta} \\ \small $\to$  RoBERTa-wwm-ext-large \end{tabular}          & \multirow{5}{*}{ \begin{tabular}[c]{@{}l@{}}24 layers \\1024-d hidden Size \\16 attention heads \end{tabular}} &\url{https://github.com/ymcui/Chinese-BERT-wwm} & \multirow{5}{*}{325M}  \cr 
\begin{tabular}[c]{@{}l@{}} MACBERT \cite{cui2020revisiting} \\ \small $\to$  MacBERT-large \end{tabular}        & & \url{https://github.com/ymcui/MacBERT} & \cr
\begin{tabular}[c]{@{}l@{}} StructBERT \cite{wang2019structbert} \\ \small $\to$  Structbert.ch.large\end{tabular}      & & \url{https://github.com/alibaba/AliceMind/tree/main/StructBERT} & \cr
\bottomrule 
\end{tabular}

\caption{Detailed information of PLMs. Models with mark $\dagger$ are utilized as the backbone in correction task. The names after $\to$ are the specific version of the PLM models.}

\label{tab:appendix-plms}
\end{table*}

%% file: Figures/tagger-label.tex
\begin{figure}[ht]
	\centering
	\includegraphics[width=\columnwidth]{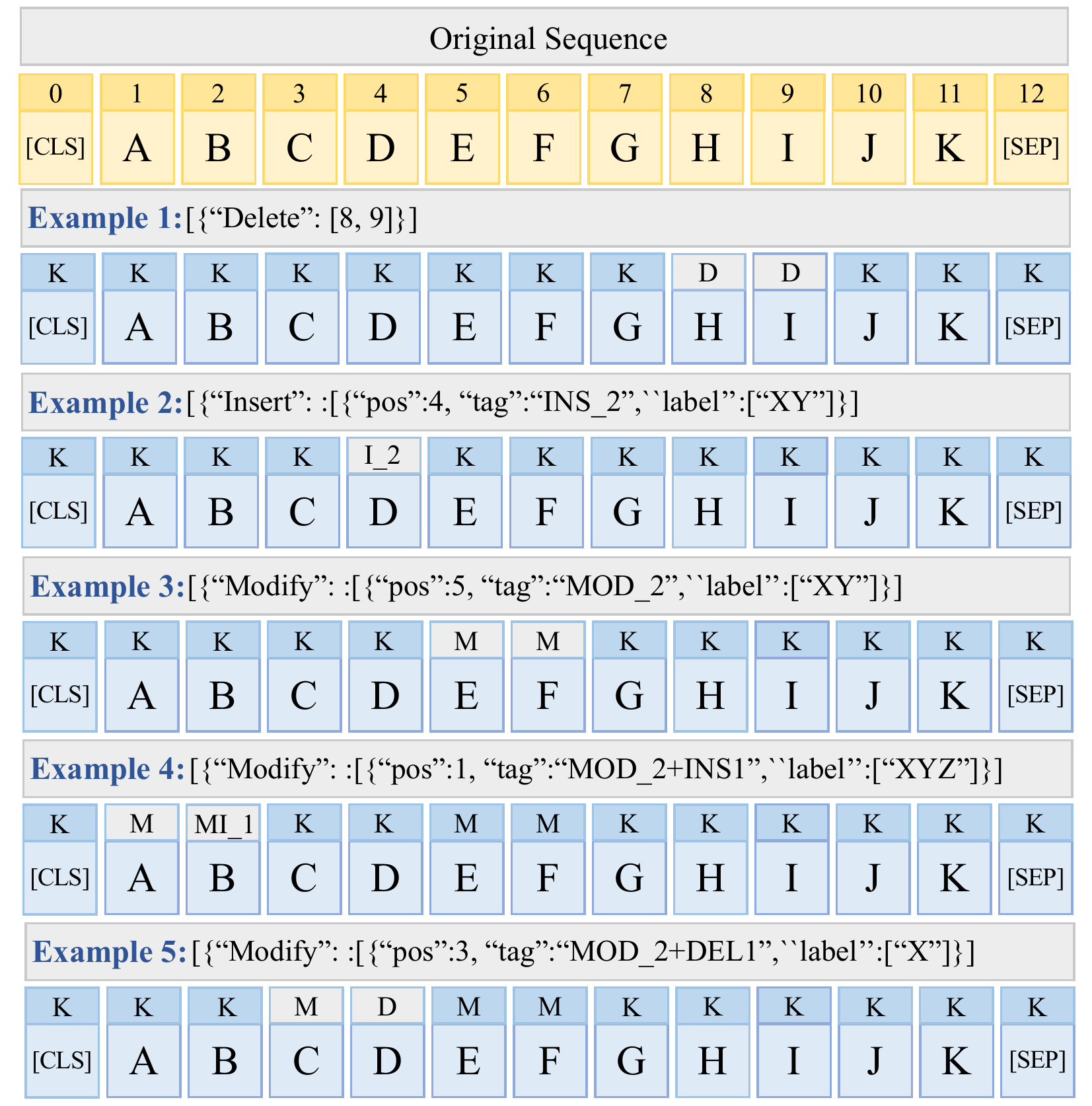} 
	\caption{Examples of the tags for five typical cases in \emph{Tagger} module.}
	\label{fig:tagger-label}
\end{figure}

%% file: Figures/generator-label.tex
\begin{figure}[t]
	\centering
	\includegraphics[width=\columnwidth]{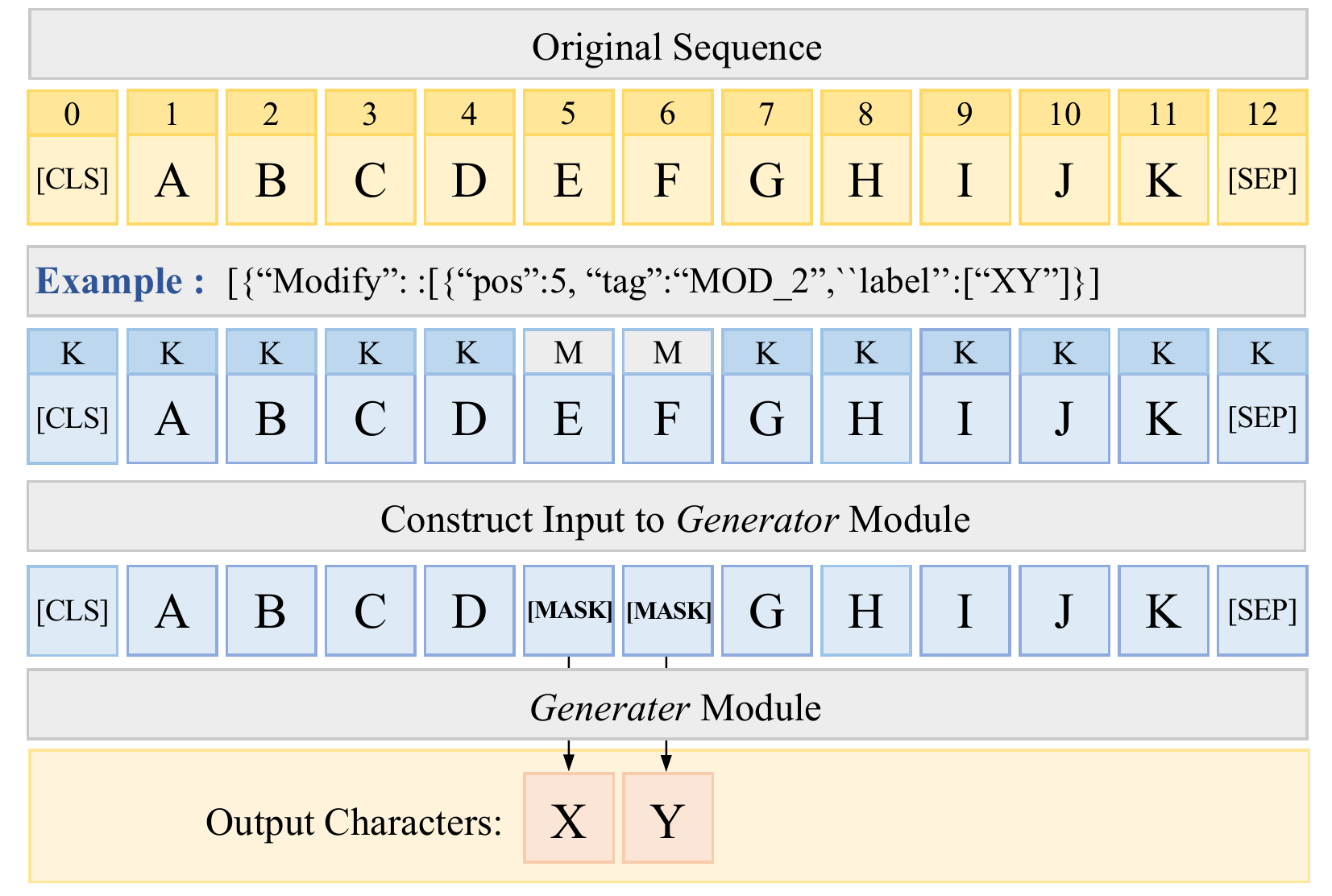} 
	\caption{An example about the model input of \emph{generator} module.}
	\label{fig:generator-label}
\end{figure}

%% file: Tables/perform-operation.tex
\begin{table*}[t]
\fontsize{10}{12}\selectfont
\centering
\begin{tabular}{p{3.5cm} p{0.88cm}<{\centering} p{0.88cm}<{\centering} p{0.88cm}<{\centering}  p{0.88cm}<{\centering}p{0.88cm}<{\centering}p{0.88cm}<{\centering}p{0.88cm}<{\centering}p{0.88cm}<{\centering}}
		\toprule
		\multirow{2}{*}{\textbf{Model}}
	     & \multicolumn{2}{c}{\textbf{Switch}} 
	     & \multicolumn{2}{c}{\textbf{Delete}} 
	     & \multicolumn{2}{c}{\textbf{Insert}} 
	     & \multicolumn{2}{c}{\textbf{Modify}}  \cr \cline{2-9}\rule[0pt]{0pt}{12pt}
		 &\textbf{EM}&\textbf{F$_{0.5}$}&\textbf{EM}&\textbf{F$_{0.5}$}&\textbf{EM}&\textbf{F$_{0.5}$}&\textbf{EM}&\textbf{F$_{0.5}$}\cr
		\midrule
		\textbf{Ratio (\%)}        &\multicolumn{2}{c}{20.90}    &\multicolumn{2}{c}{34.42}    &\multicolumn{2}{c}{27.32}    &\multicolumn{2}{c}{17.35} \cr \hdashline
		Seq2Seq (MuCGEC)           &12.33         &20.85         &31.45         &54.42         &10.90         &18.57         &8.58          &23.56\cr 
		Seq2Edit (LaserTagger)     &16.16         &20.51         &40.43         &\textbf{59.62}&7.34          &13.23         &15.18         &28.14\cr
		\textbf{STG-Joint (Ours)}  &\textbf{35.62}&\textbf{48.94}&\textbf{42.60}&59.29         &\textbf{12.79}&\textbf{18.99}&\textbf{19.14}&\textbf{34.15}\cr 
	\bottomrule
	\end{tabular}

\caption{The metric (\%) of \emph{Exact Match} and \emph{F$_{0.5}$ score} for each operation on the subset of test set. The first row (i.e., Ratio) represents the proportion of each operation to the total.}

\label{tab:perform-operate}
\end{table*}

%% file: Tables/appendix-mucgec.tex
\begin{table}[t]
    \fontsize{10}{12}\selectfont
	\centering
	\begin{tabular}{p{3.6cm}  p{0.7cm}<{\centering} p{0.7cm}<{\centering}  p{0.7cm}<{\centering}}
		\toprule
	    \textbf{Model}  & \textbf{P} & \textbf{R} & \textbf{F$_{0.5}$} \cr
		\midrule
		Seq2Seq+CPT-B           &39.47&26.19&35.84\cr 
		Seq2Seq+BART-L          &38.59&36.52&38.16\cr  \hdashline 
		Seq2Edit+RoBERTa-B      &31.06&18.74&27.45\cr
		Seq2Edit+StructBERT-L   &30.68&21.65&28.32\cr
	\bottomrule
	\end{tabular}
	\caption{Performance comparison for different PLM backbones for models in MuCGEC. The suffix of B represents base size, while L stands for large size.}
	\label{tab:appendix-mucegec}
\end{table}

%% file: Tables/appendix-valid.tex
\begin{table}[t]
    \fontsize{10}{12}\selectfont
	\centering
	\begin{tabular}{p{2.8cm} p{0.7cm}<{\centering} p{0.7cm}<{\centering} p{0.7cm}<{\centering}  p{0.7cm}<{\centering}}
		\toprule
	    \textbf{Model} & \textbf{Acc} & \textbf{P} & \textbf{R} & \textbf{F$_1$} \cr
		\midrule
		\multicolumn{5}{l}{$\bullet$ \textbf{Grammatical Error Detection}} \cr
		StructBERT-Large   &80.40&80.24&78.55&80.24\cr 
		\multicolumn{5}{l}{$\bullet$ \textbf{Grammatical Error Type Identification}} \cr
		RoBERTa-Large      &60.44&68.17&64.11&65.77\cr  \toprule
		\textbf{Model}     & \textbf{EM} & \textbf{P} & \textbf{R} & \textbf{F$_{0.5}$} \cr \midrule
		\multicolumn{5}{l}{$\bullet$ \textbf{Grammatical Error Correction}} \cr
		STG-Joint          &36.71&50.00&39.21&47.39\cr
		\bottomrule

	\end{tabular}
	\caption{Corresponding validation performance for reported test result of the best performing models on three tasks.}
	\label{tab:appendix-valid}
\end{table}

%% file: Tables/appendix-computation.tex
\begin{table}[ht]
\fontsize{9}{12}\selectfont
\begin{tabular}{p{2.6cm} p{4.2cm}}
\toprule
\textbf{Configuration} & \textbf{Value} \\ \midrule
Device                    & 1 GeForce RTX 3090 (24G RAM) \cr
PLM model                 & RoBERTa-Base-wwm-ext \cite{liu2019roberta} \cr  
Number of epochs          & 100      \cr 
Batch size                & 32 \cr
Beam size                 & 5 / [1, 5, 10, 20]\cr
Sequence Length           & 150 / [50, 100, 150]\cr
Learning Rate             & 1e-5 / [5e-6, 1e-5, 2e-5, 5e-5] \cr
Optimizer                 & Adam \cr
Dropout                   & 0.1\cr
Weight Decay              & 1e-2 \cr
Total training time       & About 12 hours \cr
Hyperparameters Selection & Best performance on the validation set with the minimum loss \cr
\bottomrule 
\end{tabular}

\caption{Computing device, hyperparamters and training time for STG-Joint model. For hyperparameters of beam size, sequence length and learning rate, the left side of / is the best hyperparameter while the right side is the set for hyperparameter search.}

\label{tab:appendix-compute}
\end{table}

%% file: Tables/case-study.tex
\begin{CJK}{UTF8}{gkai}
\begin{table*}[ht]
\centering
\fontsize{9}{12}\selectfont
\begin{tabular}{c p{4.1cm} p{4.1cm} p{1.5cm}<\centering p{1.5cm}<\centering  p{1.5cm}<\centering }
\toprule
\textbf{Type}&\textbf{Erroneous Sentence} & \textbf{Ground Truth} &\textbf{STG-Joint} & \textbf{Seq2Seq} & \textbf{Seq2Edit}\cr \midrule
\multirow{3}{*}{IWC}& \begin{tabular}[c]{@{}l@{}} \small \textbf{1.} 近些年来，我国全力做好癌\\症筛查、临床治疗和药品供应\\工作，努力\textcolor{c1}{减少}癌症死亡率。 \end{tabular} &\begin{tabular}[c]{@{}l@{}} \small 近些年来，我国全力做好癌症\\ 筛查、临床治疗和药品供应工\\作，努力\textcolor{c1}{降低}癌症死亡率。\end{tabular} & \CheckmarkBold & \XSolidBrush & \CheckmarkBold \cr 

& \begin{tabular}[c]{@{}l@{}} \small \textbf{2.} 我们有吃苦耐劳的人民，又\\有\textcolor{c1}{优裕}的自然资源。 \end{tabular} &\begin{tabular}[c]{@{}l@{}} \small 我们有吃苦耐劳的人民，又有\\\textcolor{c1}{丰富}的自然资源。 \end{tabular} & \XSolidBrush & \XSolidBrush & \XSolidBrush \rule[0pt]{0pt}{12pt} \cr \hline

\multirow{4}{*}{CM}& \begin{tabular}[c]{@{}l@{}} \small \textbf{3.} 笔记本电脑充分显示了快捷\\、稳定、方便而成为各种赛事\\新闻报道的重要工具。 \end{tabular} &\begin{tabular}[c]{@{}l@{}} \small 笔记本电脑充分显示了快捷、\\稳定、方便\textcolor{c1}{的特点}而成为各种\\赛事新闻报道的重要工具。 \end{tabular} & \CheckmarkBold & \XSolidBrush & \XSolidBrush \rule[0pt]{0pt}{15pt}\cr 

& \begin{tabular}[c]{@{}l@{}} \small \textbf{4.} 我们要养成爱读书，特别是\\读经典，读名著，让书香溢满\\校园。 \end{tabular} &\begin{tabular}[c]{@{}l@{}} \small 我们要养成爱读书，特别是读\\经典，读名著\textcolor{c1}{的习惯}，让书香\\溢满校园。 \end{tabular} & \XSolidBrush & \XSolidBrush & \XSolidBrush \rule[0pt]{0pt}{12pt} \cr \hline

\multirow{3}{*}{CR}& \begin{tabular}[c]{@{}l@{}} \small \textbf{5.} 为精简字数，这篇文章不得\\不略加删改\textcolor{c1}{一些}。 \end{tabular} &\begin{tabular}[c]{@{}l@{}} \small 为精简字数，这篇文章不得不\\略加删改。\end{tabular} & \CheckmarkBold & \CheckmarkBold & \CheckmarkBold \rule[0pt]{0pt}{15pt}\cr 

& \begin{tabular}[c]{@{}l@{}} \small \textbf{6.} 我们发自内心\textcolor{c1}{由衷}地感谢老\\师多年来的默默付出。 \end{tabular} &\begin{tabular}[c]{@{}l@{}} \small 我们发自内心地感谢老师多年\\来的默默付出。 \end{tabular} & \CheckmarkBold & \XSolidBrush & \XSolidBrush \rule[0pt]{0pt}{12pt} \cr \hline

\multirow{5}{*}{SC}& \begin{tabular}[c]{@{}l@{}} \small \textbf{7.} 一个人变好变坏，\textcolor{c1}{关键在于}\\内因起决定作用。 \end{tabular} &\begin{tabular}[c]{@{}l@{}} \small 一个人变好变坏，内因起决定\\作用。\end{tabular} & \CheckmarkBold & \CheckmarkBold & \CheckmarkBold \rule[0pt]{0pt}{15pt}\cr 

& \begin{tabular}[c]{@{}l@{}} \small \textbf{8.} 期末考试前出现失眠、烦躁\\等现象，这往往是\textcolor{c1}{因为}太在乎\\考试成绩，心理负担过重造成\\的。 \end{tabular} &\begin{tabular}[c]{@{}l@{}} \small 期末考试前出现失眠、烦躁等\\现象，这往往是太在乎考试成\\绩，心理负担过重造成的。 \end{tabular} & \CheckmarkBold & \XSolidBrush & \XSolidBrush \rule[0pt]{0pt}{12pt} \cr \hline

\multirow{6}{*}{IWO}& \begin{tabular}[c]{@{}l@{}} \small \textbf{9.} 参加奥运的选手们十分清楚\\，一场比赛的输赢，不仅关系\\到\textcolor{c1}{祖国的荣誉}，而且关系到\textcolor{c1}{个}\\\textcolor{c1}{人的尊严}。\end{tabular} &\begin{tabular}[c]{@{}l@{}} \small 参加奥运的选手们十分清楚，\\一场比赛的输赢，不仅关系到\\\textcolor{c1}{个人的尊严}，而且关系到\textcolor{c1}{祖国}\\\textcolor{c1}{的荣誉}。\end{tabular} & \CheckmarkBold & \XSolidBrush & \XSolidBrush \rule[0pt]{0pt}{15pt}\cr 

& \begin{tabular}[c]{@{}l@{}} \small \textbf{10.}\,\,\, 学校\textcolor{c1}{自从}开展研究性学习\\以来，同学们踊跃参与，创新\\意识和创新能力得到很大的提\\升。
\end{tabular} &\begin{tabular}[c]{@{}l@{}} \small \textcolor{c1}{自从}学校开展研究性学习以来\\，同学们踊跃参与，创新意识\\和创新能力得到很大的提升。\end{tabular} & \CheckmarkBold & \CheckmarkBold & \CheckmarkBold \rule[0pt]{0pt}{12pt} \cr \hline

\multirow{4}{*}{ILL}& \begin{tabular}[c]{@{}l@{}} \small \textbf{11.}\,\,\, “十一”放假之前，老师反\\复强调要防止\textcolor{c1}{不}发生事故。\end{tabular} &\begin{tabular}[c]{@{}l@{}} \small “十一”放假之前，老师反复强\\调要防止发生事故。\end{tabular} & \CheckmarkBold & \CheckmarkBold & \CheckmarkBold \rule[0pt]{0pt}{15pt}\cr 

& \begin{tabular}[c]{@{}l@{}} \small \textbf{12.}\,\,\, 面对一件棘手的事情，我\\们需要三思而后行。多动脑子\\，会\textcolor{c1}{避免}少捅娄子少出错。\end{tabular} &\begin{tabular}[c]{@{}l@{}} \small 面对一件棘手的事情，我们需\\要三思而后行。多动脑子，会\\少捅娄子少出错。\end{tabular} & \XSolidBrush & \XSolidBrush & \XSolidBrush \rule[0pt]{0pt}{15pt}\cr \hline

\multirow{6}{*}{AM}& \begin{tabular}[c]{@{}l@{}} \small \textbf{13.}\,\,\, 很多人认为科学家终日埋\\头搞科研，不问家事，有点儿\\不近人情，然而事实却是对这\\种偏见的最好\textcolor{c1}{说明}。\end{tabular} &\begin{tabular}[c]{@{}l@{}} \small 很多人认为科学家终日埋头搞\\科研，不问家事，有点儿不近\\人情，然而事实却是对这种偏\\见的最好反驳。\end{tabular} & \CheckmarkBold & \XSolidBrush & \XSolidBrush \rule[0pt]{0pt}{15pt}\cr

& \begin{tabular}[c]{@{}l@{}} \small \textbf{14.}\,\,\, 他决定\textcolor{c1}{背}着妈妈去医院检\\查身体。\end{tabular} &\begin{tabular}[c]{@{}l@{}} \small 他决定\textcolor{c1}{瞒}着妈妈去医院检查身\\体。
\end{tabular} & \XSolidBrush & \XSolidBrush & \XSolidBrush \rule[0pt]{0pt}{15pt}\cr

& \begin{tabular}[c]{@{}l@{}} \small \textbf{15.}\,\,\, 张义和王强上课说话，被\\老师叫去办公室了。
\end{tabular} &\begin{tabular}[c]{@{}l@{}} \small 张义和王强上课说话，\textcolor{c1}{两人}被\\老师叫去办公室了。\end{tabular} & \XSolidBrush & \XSolidBrush & \XSolidBrush \rule[0pt]{0pt}{12pt} \cr 

\bottomrule 
\end{tabular}

\caption{The case study for comparing the performances of models. The characters in red denote the differences between erroneous sentences and ground truth. We demonstrate the English version in Table~\ref{tab:case-study-eng}.}

\label{tab:case-study}
\end{table*}
\end{CJK}

%% file: Tables/case-study-eng.tex
\begin{CJK}{UTF8}{gkai}
\begin{table*}[ht]
\centering
\fontsize{9}{12}\selectfont
\begin{tabular}{c p{5cm} p{5cm} p{3.5cm}}
\toprule
\textbf{Type}&\textbf{Erroneous Sentence} & \textbf{Ground Truth} & \textbf{Tips}\cr \midrule
\multirow{7}{*}{IWC}& \textbf{1.} In recent years, we have struggled to \textcolor{c1}{cut down} cancer mortality by improving cancer screening, clinical care and the supply of drugs. & In recent years, we have struggled to \textcolor{c1}{reduce} cancer mortality by improving cancer screening, clinical care and the supply of drugs. & In general, ``cut down'' cannot be collocated with ``cancer mortality'', and ``reduce'' should be used.\cr \rule[0pt]{0pt}{10pt}

&\textbf{2.} We have industrious people and \textcolor{c1}{favourable} natural resources. & We have industrious people and \textcolor{c1}{abundant} natural resources. & Usually, we pair ``abundant'' with ``resources''.\cr \hline 

\multirow{8}{*}{CM}&  \textbf{3.} Laptops demonstrate the speed, stability and convenience and become an important tool for news coverage of various events. & Laptops demonstrate the \textcolor{c1}{characteristics of} speed, stability and convenience and become an important tool for news coverage of various events. &In Chinese, the incorrect sentence is missing the object ``characteristics''.\cr \rule[0pt]{0pt}{10pt} 

& \textbf{4.} We should develop of reading, especially reading classics and masterpieces, so that the fragrance of books can overflow the campus. & We should develop \textcolor{c1}{the habit} of reading, especially reading classics and masterpieces, so that the fragrance of books can overflow the campus. & Similarly, the erroneous sentence misses the object ``habbit'' in Chinese.\cr \hline

\multirow{5}{*}{CR}& \textbf{5.} In order to reduce the word count, this article had to be slightly redacted \textcolor{c1}{some}. & In order to reduce the word count, this article had to be slightly redacted. & The word ``some'' is redundant and can be deleted.\cr \rule[0pt]{0pt}{10pt}

& \textbf{6.} From the bottom of our hearts, we thank our teachers \textcolor{c1}{sincerely} for their quiet dedication over the years. &From the bottom of our hearts, we thank our teachers for their quiet dedication over the years. & In Chinese, the word ``sincerely'' is superfluous and should be removed. \cr \hline

\multirow{7}{*}{SC}& \textbf{7.} \textcolor{c1}{The crucial thing} for a person to become good or bad is that the inner reasons play a role in determining it. & For a person to become good or bad, the inner reasons play a role in determining it. & The structure of ``the crucial thing for'' and ``play a role in'' are confusing.\cr \rule[0pt]{0pt}{10pt}

& \textbf{8.} Insomnia and irritability before exams are caused by \textcolor{c1}{due to} the excessive psychological burden with caring too much about exam results.  &Insomnia and irritability before exams are caused by the excessive psychological burden with caring too much about exam results. & We can not apply the structure of ``caused by'' and ``due to'' in a sentence simultaneously.\cr \hline

\multirow{10}{*}{IWO}&\textbf{9.} Olympic athletes understand that winning or losing a race is not only about the \textcolor{c1}{honor of the country}, but also about the \textcolor{c1}{dignity of themselves}.& Olympic athletes understand that winning or losing a race is not only about the \textcolor{c1}{dignity of themselves}, but also about the \textcolor{c1}{honor of the country.} & ``Honor of the country'' and ``dignity of themselves'' should be swapped due to progressive relationship.\cr \rule[0pt]{0pt}{10pt}

& \textbf{10.} The school \textcolor{c1}{since} introduced the research study, students have participated enthusiastically and their creative awareness and ability have been greatly enhanced. &\textcolor{c1}{Since} the school introduced the research study, students have participated enthusiastically and their creative awareness and ability have been greatly enhanced. & ``Since'' should be placed at the beginning of the sentence\cr \hline

\multirow{6}{*}{ILL}& \textbf{11.} Before the holiday, teachers emphasized over and over again to prevent accidents from \textcolor{c1}{not} happening. &Before the holiday, teachers emphasized over and over again to prevent accidents from happening. & The double negation causes logical errors.\cr \rule[0pt]{0pt}{10pt}

& \textbf{12.} When faced with difficulties, we need to think first. More thinking will \textcolor{c1}{avoid} less troubles and mistakes. & When faced with difficulties, we need to think first. More thinking \textcolor{c1}{can lead to} less troubles and mistakes. & More thinking leads to less errors in commonsense, while ``avoid'' causes errors.\cr \hline

\multirow{9}{*}{AM}&\textbf{13.} Many people think that scientists engage in research and do not communicate with others, yet the facts are the best \textcolor{c1}{illustration of} this prejudice. &Many people think that scientists engage in research and do not communicate with others, yet the facts are the best \textcolor{c1}{rebuttal to} this prejudice. & The semantic meaning is rather ambiguous that we cannot infer the role of facts on prejudice. \cr \rule[0pt]{0pt}{10pt}

& \textbf{14.}He decided  \textcolor{c1}{to} \textcolor{c1}{carry} \textcolor{c1}{(not to tell)} his mother \textcolor{c1}{(he was going)} to the hospital. &He decided \textcolor{c1}{not to tell} his mother he was going to the hospital. & There is an ambiguity in Chinese.\cr \rule[0pt]{0pt}{10pt}

&\textbf{15.} Yi Zhang and Qiang Wang talked in class and \textcolor{c1}{was} called to the office by the teacher. &  Yi Zhang and Qiang Wang talked in class and \textcolor{c1}{the two were} called to the office by the teacher.& There is ambiguity on who was called to the office.\cr 

\bottomrule 
\end{tabular}

\caption{The English version of erroneous sentences and ground truth in case study. Furthermore, we provide tips for better understanding the grammatical errors in Chinese.}

\label{tab:case-study-eng}
\end{table*}
\end{CJK}